\documentclass[accepted]{uai2021}

\usepackage{amssymb}
\usepackage{amsthm}
\usepackage{mathtools}

\usepackage{microtype}
\usepackage{graphicx}
\usepackage{booktabs} %
\usepackage{caption}
\usepackage{subcaption}
\usepackage{multirow}
\usepackage{xcolor}

\usepackage[round]{natbib}

\usepackage{algorithm}
\usepackage[noend]{algorithmic}

\DeclareMathAlphabet{\mathcal}{OMS}{cmsy}{m}{n}

\newcommand{\U}{\mathcal U}
\newcommand{\X}{\mathcal X}
\newcommand{\Z}{\mathcal Z}
\newcommand{\R}{\mathbb R}

\newcommand{\KL}[2]{D_{\text{KL}} \!\left(#1 \mid\mid #2\right)}

\newcommand{\ELBO}{\mathcal L}
\newcommand{\ELBOCIF}{\ELBO_2}
\newcommand{\ELBONF}{\ELBO_1}
\DeclareMathOperator{\E}{\mathbb E}
\newcommand{\covmat}{\Sigma}

\newcommand{\mytitle}{Variational Inference with Continuously-Indexed Normalizing Flows}

\newtheorem{theorem}{Theorem}[section]

\newtheorem{proposition}[theorem]{Proposition}

\title{\mytitle}

\author[1]{\href{mailto:Anthony Caterini <anthony.caterini@stats.ox.ac.uk>?Subject=Your UAI 2021 paper}{Anthony~Caterini}{}} %
\author[1]{Rob~Cornish}
\author[1]{Dino~Sejdinovic}
\author[1]{Arnaud~Doucet}
\affil[1]{Department of Statistics, University of Oxford}

\begin{document}
\maketitle

\begin{abstract} 
Continuously-indexed flows (CIFs) have recently achieved improvements over baseline normalizing flows on a variety of density estimation tasks.
CIFs do not possess a closed-form marginal density, and so, unlike standard flows, cannot be plugged in directly to a variational inference (VI) scheme in order to produce a more expressive family of approximate posteriors.
However, we show here how CIFs can be used as part of an \emph{auxiliary} VI scheme to formulate and train expressive posterior approximations in a natural way.
We exploit the conditional independence structure of multi-layer CIFs to build the required auxiliary inference models, which we show empirically yield low-variance estimators of the model evidence.
We then demonstrate the advantages of CIFs over baseline flows in VI problems when the posterior distribution of interest possesses a complicated topology, obtaining improved results in both the Bayesian inference and surrogate maximum likelihood settings.
\end{abstract}

\section{Introduction} \label{sec:intro}
Variational inference (VI) has emerged as a fast, albeit biased, alternative to Markov chain Monte Carlo for Bayesian inference.
VI methods attempt to minimize the KL divergence from a parametrized family of distributions to a true posterior over latent variables.
The expressiveness of this family is essential for good performance, with under-expressive models leading to both increased bias and underestimation of posterior variance \citep{yin2018semi}.

If the density of the approximate posterior is available in closed-form, then the variational family is said to be \emph{explicit}.
Explicit models allow for straightforward estimation of the VI objective, but can often lead to reduced expressiveness, which limits their performance overall.
Mean-field VI \citep{blei2017variational}, for example, imposes restrictive independence assumptions between the latent variables of interest.

Normalizing flows \citep{tabak2010density, rezende2015variational} provide an alternative family of explicit density models that yield improved expressiveness compared with mean field alternatives.
These methods push samples from a simple base distribution (typically Gaussian) through parametrized bijections to produce complex, yet still exact, density models.
Normalizing flows have performed well in tasks requiring explicit density models (e.g.\ \citep{louizos2017multiplicative, papamakarios2017masked, ho2019compression}), including VI, where flows have demonstrated the ability to improve the quality of approximate posteriors \citep{rezende2015variational, durkan2019neural}.

Although normalizing flows can directly improve the expressiveness of mean-field VI schemes, their inherent bijectivity remains quite restrictive.
We can overcome this limitation by instead using continuously-indexed flows (CIFs) \citep{cornish2019relaxing}.
CIFs relax the bijectivity constraint of standard normalizing flows by augmenting them with continuous index variables, thus parametrizing an \emph{implicit} density model defined as the marginalization over these additional indexing variables.
Beyond being well-grounded theoretically, CIFs also have empirically demonstrated the ability to outperform relevant normalizing flow baselines in the context of density estimation, and thus it is sensible to investigate the performance of CIFs in VI.

A difficulty in applying CIFs to VI -- and implicit models more generally -- is that their marginal distribution is intractable, precluding evaluation of the standard VI objective.
However, conveniently, CIFs still admit a tractable \emph{joint} distribution over the variables of interest (latent variables in VI) and the auxiliary indexing variables.
We can therefore appeal to the framework of \emph{auxiliary variational inference} (AVI) \citep{agakov2004auxiliary}, which facilitates the training of implicit models with tractable joint densities as variational inference models.
CIFs also \emph{already} prescribe a model for inferring auxiliary variables -- typically required in AVI schemes -- suggesting that CIFs are a natural fit here.
AVI methods more generally have shown improved expressiveness over the explicit counterparts in several settings \citep{DBLP:journals/corr/BurdaGS15, yin2018semi}, and are becoming more popular with the rise of implicit models overall \citep{tran2017hierarchical, lawson2019energy, kleinegesse2020sequential}, suggesting that this framework is able to overcome any supposed drawbacks associated with not having access to explicit densities.

In this work, we show that these benefits are also realized when CIFs are applied within the AVI framework.
We first describe how CIFs can be used as the variational family in AVI, naturally incorporating the components of CIF models designed for density estimation, and we explain how we can also  \emph{amortize} these inference models.
We then empirically demonstrate the advantages of using CIFs over standard normalizing flows for modelling posteriors with complicated topologies, and additionally how CIFs can facilitate maximum likelihood estimation of the parameters of complex latent-variable generative models.

\section{Continuously-Indexed Flows for Variational Inference}

In this section we first review necessary background on variational inference (VI) -- including auxiliary variational inference (AVI) -- and continuously-indexed flows (CIFs).
We then describe how CIFs naturally fit in as a class of auxiliary variational posteriors, and extend to include amortization.
We summarize the results of this section in \autoref{alg:ELBO}.

\subsection{VARIATIONAL INFERENCE}

Given a joint probability density $p_{X,Z}$, with observed data $X \in \X$ and latent variable $Z \in \Z$, variational inference (VI) provides us with a means to approximate the intractable posterior $p_{Z|X}(\cdot \mid x)$.
This is accomplished by introducing a parametrized approximate posterior\footnote{We may also \emph{amortize} $q_Z$ and replace it with the conditional $q_{Z|X}$, especially when using VI to facilitate generative modelling.
Further discussion on amortization is deferred to \autoref{sec:amortization-main}.} $q_Z$, and maximizing the evidence lower bound (ELBO)
\begin{equation} \label{eq:elbo_1}
  \ELBO_1(x) \coloneqq \E_{z \sim q_Z} [ \log p_{X, Z}(x, z) - \log q_Z(z) ]
\end{equation}
with respect to the parameters of $q_Z$.
This is equivalent to minimizing the KL divergence between $q_Z$ and the true posterior $p_{Z|X}(\cdot \mid x)$.

Explicit VI methods, such as mean-field approaches or normalizing flow models, define $q_Z$ in such a way that it can be evaluated pointwise.
Although this approach is computationally convenient, the expressiveness of the resulting methods can often be limited.
To improve on this, implicit methods define $q_Z$ typically through some type of sampling process with intractable marginal distribution, such as the pushforward of a simple distribution through an unrestricted deep neural network.
These methods can be quite powerful but also challenging to optimize, especially in the context of VI \citep{tran2017hierarchical}, as we lose the tractability of \eqref{eq:elbo_1}.

\paragraph{Auxiliary Variational Inference} In contexts where $q_Z$ is obtained as $q_Z(z) \coloneqq \int q_{Z,U}(z, u) \, \mathrm{d}u$ for some joint density $q_{Z,U}$ that can be sampled from and evaluated pointwise, its parameters can be learned via \emph{auxiliary variational inference} (AVI) \citep{agakov2004auxiliary}.
We refer to $U$ here as an \emph{auxiliary} variable.
These approaches introduce an auxiliary inference distribution $r_{U\mid Z}$ and optimize
\begin{equation} \label{eq:elbo_2}
    \ELBO_2(x) \!\coloneqq\! \E_{(z, u) \sim q_{Z, U}}\!\left[ \log \frac{ p_{X, Z}(x, z) \cdot r_{U|Z}(u \mid z)}{q_{Z, U}(z, u)} \right]\!.
\end{equation}
Key to this approach is the fact that $\ELBO_1(x)\! \geq\! \ELBO_2(x)$, and that this bound is tight when $r_{U\mid Z} = q_{U\mid Z}$, which holds because
\begin{equation} \label{eq:elbo_difference}
    \ELBO_1 (x) \!= \!\ELBO_2(x)\! +\! \E_{z \sim q_Z}\! \left[ D_\text{KL}(q_{U|Z}(\cdot | z) || r_{U|Z}(\cdot | z)) \right].
\end{equation}
As such, optimizing the parameters of $r_{U \mid Z}$ jointly with those of $q_{Z,U}$ will encourage learning better approximations to the true posterior $p_{Z\mid X}$.
Note that, although we now are optimizing a lower bound on $\mathcal L_1$, we are also optimizing over a larger family of approximate posteriors which may end up yielding a better optimum (cf.\ \autoref{prop:cif_vs_baseline} below).

\subsection{CONTINUOUSLY-INDEXED FLOWS}

We now describe in detail the \emph{continuously-indexed flow} (CIF) model \citep{cornish2019relaxing}, which we intend to incorporate into an AVI scheme.
CIFs define a density $q_Z$ over $\Z$ as the $Z$-marginal of
\begin{equation} \label{eq:cif_generative}
    W \sim q_W, \quad U \sim q_{U | W}(\cdot \mid W), \quad Z = G(W; U),
\end{equation}
where $q_W$ is a noise distribution over $\Z$, $q_{U \mid W}$ is a conditional distribution over $\U$ describing an auxiliary indexing variable, and $G : \Z \times \U \rightarrow \Z$ is a function such that $G(\cdot; u)$ is a bijection for each $u \in \U$.
For all $z \in \Z$, the density model $q_Z$ is then given by the intractable integral $q_Z(z) \coloneqq \int q_{Z,U}(z, u) \, \mathrm d u$ over the tractable joint density $q_{Z,U}$ given by
\begin{align} \label{eq:cif_joint}
    &q_{Z, U} (z, u) = q_W\!\left(G^{-1}(z; u)\right) \\
    &\quad \times q_{U|W}\!\left(u \mid G^{-1}(z; u)\right) \nonumber \left|\det \mathrm D_z G^{-1}(z; u)\right|
\end{align}
for all $z \in \Z$ and $u \in \U$, where $G^{-1}$ denotes the inverse of $G$ (and $\mathrm D_z G^{-1}$ the Jacobian of $G^{-1}$) with respect to its first argument $z$ (see \autoref{sec:app-density-obj} for a derivation).
Typically, $q_{U|W}$ is chosen to be conditionally Gaussian with mean and covariance as the outputs of neural networks taking the conditioning variables $W$ and $Z$ as input, and
\begin{equation} \label{eq:cif_G}
    G(w; u) \coloneqq e^{s(u)} \odot \left(g(w) + t(u)\right),
\end{equation}
where $g : \Z \rightarrow \Z$ is some base bijection,  $s, t: \U \rightarrow Z$ are arbitrary neural networks, and $\odot$ denotes elementwise multiplication.
\citet{cornish2019relaxing} used the model \eqref{eq:cif_generative} in the context of density estimation to model the generative process of a set of i.i.d.\ data.

\paragraph{Multi-layer CIFs} \citet{cornish2019relaxing} also propose to improve the expressiveness of \eqref{eq:cif_generative} by taking the noise distribution $q_W$ to be a CIF model itself.
Applying this recursively $L$ times, we can take $q_Z$ to be the $W_L$-marginal in the following model:
\begin{align} \label{eq:cif_generative_multilayer}
    W_0 \sim q_{W_0}, \quad U_\ell &\sim q_{U_\ell | W_{\ell-1}}(\cdot \mid W_{\ell-1}) \nonumber \\
    W_\ell &= G_\ell(W_{\ell-1}; U_\ell), 
\end{align}
where $\ell \in \{1, \ldots, L\}$.
Here $q_{W_0}$ is typically a mean-field Gaussian and each $G_\ell : \mathcal Z \times \mathcal U \rightarrow \mathcal Z$ is bijective in its first argument.
Practically, multi-layer CIF models have demonstrated far more representational power than single-layer versions, although we note that we can still view this multi-layer model as an instance of \eqref{eq:cif_generative} for certain choices of $q_{U|W}$ and $G$ (as in \autoref{sec:app-stacking}).%

\paragraph{Auxiliary Inference Distribution} The intractability of $q_Z$ arising from both \eqref{eq:cif_generative} and \eqref{eq:cif_generative_multilayer} precludes direct maximum likelihood estimation.
\citet{cornish2019relaxing} therefore introduce an auxiliary \emph{backward} distribution, either $r_{U|Z}$ or $r_{U_{1:L}|Z}$ respectively, to enable training of CIFs through an amortized ELBO.
Particularly noteworthy is the structure of this distribution in the multi-layer case. 
The optimal choice for $r_{U_{1:L}|Z}$ would be $q_{U_{1:L}|Z}$, which can be shown to factorize as
$q_{U_{1:L} | Z}(u_{1:L} \mid z) = \prod_{\ell=1}^L q_{U_\ell | W_\ell}(u_\ell \mid w_\ell),$
where $w_L \coloneqq z$ and $w_\ell \coloneqq G^{-1}_{\ell+1}(w_{\ell+1}; u_{\ell+1})$ recursively for $\ell \in \{1, \ldots, L-1\}$.
Although this gives us the form of $q_{U_{1:L} | Z}$, the backward distributions $q_{U_\ell | W_\ell}$ are not generally available in closed form.
However this does at least motivate defining $r_{U_{1:L} | Z}$ to have the same form, which can be done by introducing (reparametrizable) densities $r_{U_\ell | W_\ell}$ and setting
\begin{equation} \label{eq:multilayer_backward}
    r_{U_{1:L} | Z}(u_{1:L} \mid z) \coloneqq \prod_{\ell=1}^L r_{U_\ell | W_\ell}(u_\ell \mid w_\ell)
\end{equation}
with $w_\ell$ defined as above.
The densities for $r_{U_\ell | W_\ell}$ are also taken to be parametrized conditional Gaussians.
This structured inference procedure induces a natural weight-sharing scheme between the forward and backward directions of the model, as both are defined using $G_\ell$.

\subsection{CIF MODELS IN AVI} \label{sec:cif-avi}

We can use CIFs as the family of approximate posteriors $q_Z$ in VI by appealing to the framework of AVI.
Starting with the single-layer version, we see from \eqref{eq:cif_joint} that CIFs admit a tractable joint distribution $q_{Z,U}$ over latent and auxiliary variables.
We can then plug this distribution into \eqref{eq:elbo_2}, noting also that CIFs already prescribe a form for $r_{U|Z}$ and thus are a natural fit within an AVI scheme.
However, we must take one additional step to formulate an objective amenable to optimization, as na\"ively substituting $q_{Z,U}$ into \eqref{eq:elbo_2} produces an expectation over a distribution containing the parameters of $G$ itself.
To address this, we show in \autoref{sec:app-density-obj} how to rewrite this as an expectation over $q_{W,U}$ rather than $q_{Z,U}$, obtaining the objective
\begin{align} \label{eq:elbo_cif}
    &\E_{(w, u) \sim q_{W,U}}\! \!\left[ \log \frac{p_{X, Z}(x, z)\! \cdot\! r_{U|Z}(u \mid z)}{q_{W,U}(w,u) \!\cdot \!|\det \mathrm D_w G(w; u) |^{\!-\!1}}\right],
\end{align}
where we write $z \coloneqq G(w; u)$ for readability.
We always select $q_{W,U}$ to be reparametrizable \citep{DBLP:journals/corr/KingmaW13}, which makes the objective straightforward to optimize via stochastic gradient descent with respect to the parameters of $q, r,$ and $G$.
Note however that $r_{U\mid Z}$ need not necessarily be reparametrizable.
This ``direction'' of reparametrization contrasts with CIF models for density estimation which require $r_{U|Z}$ -- not $q_{U|W}$ -- to be reparametrizable.
Further discussion demonstrating that CIF models in density estimation and VI can be viewed as ``opposites'' of each other is provided in \autoref{sec:cif_for_de} and \autoref{sec:app-cif_de_vi}.

\begin{figure*} 
\centering
\begin{subfigure}{.49\textwidth}
  \centering
  \includegraphics[width=.95\linewidth]{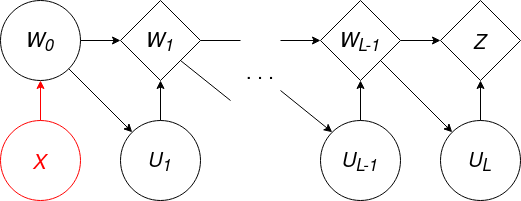}
  \caption{Sampling $Z \sim q_{Z \textcolor{red}{| X}}$ as defined in \eqref{eq:cif_generative}}
  \label{fig:sub-first}
\end{subfigure}
\begin{subfigure}{.49\textwidth}
  \centering
  \includegraphics[width=.95\linewidth]{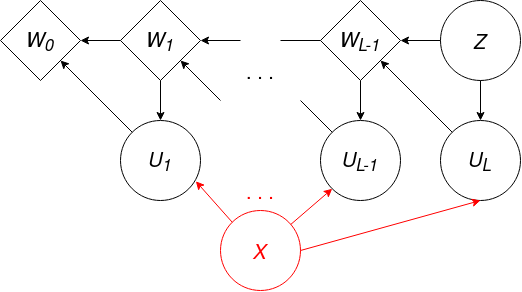}
  \caption{Sampling $U_{1:L} \sim r_{U_{1:L} | Z, \textcolor{red}{X}}$ as defined in \eqref{eq:multilayer_backward}}
  \label{fig:sub-second}
\end{subfigure}
\caption{Diagrams demonstrating how to sample from the CIF approximate posterior (left) and the auxiliary inference model (right).
The \textcolor{red}{red highlighting} corresponds to \textcolor{red}{amortization} -- these can be ignored for models not requiring amortization.}
\label{fig:model_schematic}
\end{figure*}

\paragraph{Multi-layer CIFs in AVI} We can also use multi-layer CIFs as past of an AVI scheme. 
Now, each of the $q_{U_\ell | W_{\ell-1}}$ distributions in \eqref{eq:cif_generative_multilayer} is chosen to be reparametrizable, again contrasting with \citep{cornish2019relaxing} which does not require reparametrizable distributions here.
\autoref{fig:sub-first} graphically displays the joint model $q_{Z,U_{1:L}}$.
We also adopt the form of $r_{U_{1:L}|Z}$ from \eqref{eq:multilayer_backward} and demonstrate this auxiliary inference procedure in \autoref{fig:sub-second}, although we do not require the individual $r_{U_\ell | W_\ell}$ distributions to be reparametrizable.
Being able to have $r_{U_{1:L}|Z}$ match the structure of the true auxiliary posterior $q_{U_{1:L}|Z}$ is likely useful in lowering the variance of estimators of the ELBO and gradients thereof.

We can now substitute our definitions for $q_{Z, U_{1:L}}$ (implied by \eqref{eq:cif_generative_multilayer}) and $r_{U_{1:L} \mid Z}$ into \eqref{eq:elbo_2} to derive an optimization objective for training multi-layer CIFs as the approximate posterior in VI.
We again must be careful about reparametrization, as we need to write the objective as an expectation over $q_{W_0, U_{1:L}}$ instead of $q_{Z, U_{1:L}}$ to be able optimize all parameters of $q$, analogously to \eqref{eq:elbo_cif}.
Further details on how to do this are provided in  \autoref{sec:app-stacking}, with the full objective given in \eqref{eq:cif_multilayer_obj}.
\autoref{alg:ELBO} describes how to compute an unbiased estimator of this objective, from which we can then obtain unbiased gradients via automatic differentiation.

\subsection{Amortization} \label{sec:amortization-main}

VI methods can also be used to provide a surrogate objective for maximum likelihood estimation of the parameters of latent-variable models, particularly for deep generative models such as the variational auto-encoder (VAE) \citep{DBLP:journals/corr/KingmaW13, rezende2014stochastic}.
In these settings, the goal is to maximize the marginal log-likelihood $\sum_i \log p_X(x_i)$ over the observed data $\{x_i\}_i$, with respect to the parameters of $p$, where $\log p_X(x) \coloneqq \log \int p_{X,Z}(x,z) \,\mathrm d z$ is a density model containing parametrized $p_{X,Z}(x,z)$.
This integral is often intractable, and thus we resort to maximizing the ELBO \eqref{eq:elbo_1} (or \eqref{eq:elbo_2}) with respect to both the parameters of $p$ and $q$ as it bounds the marginal log-likelihood from below.
In this case, we would like to \emph{amortize} the cost of variational inference across an entire dataset, rather than compute a brand new approximate posterior for each datapoint, and so we parametrize our variational distribution as an explicit function of the data.

We can readily incorporate amortization into the single-layer CIF by replacing $q_W$ with $q_{W|X}$ in \eqref{eq:cif_generative}, and $r_{U|Z}$ with $r_{U|Z, X}$ in \eqref{eq:elbo_cif} since the true auxiliary posterior $q_{U|Z, X}$ will now carry an explicit dependence on the data $X$.
For multi-layer CIFs, it is again straightforward to incorporate amortization into the model for $q$ by replacing $q_{W_0}$ with $q_{W_0 | X}$ in \eqref{eq:cif_generative_multilayer}.
Additional care must be taken when constructing the \emph{auxiliary} inference model $r$, however, as the explicit dependence on data will appear in each term of the factorization of the true auxiliary posterior $q_{U_{1:L} | Z, X}$.
We thus structure $r_{U_{1:L} | Z, X}$ similarly:
\[
    r_{U_{1:L} | Z, X}(u_{1:L} \mid z, x) \coloneqq \prod_{\ell=1}^L r_{U_\ell | W_\ell, X}(u_\ell \mid w_\ell, x),
\]
where $w_\ell$ is as defined in \eqref{eq:multilayer_backward}.
The full amortized objective is given in \eqref{eq:a_cif_multilayer_obj} in the Appendix.
\autoref{fig:model_schematic} graphically demonstrates how to incorporate amortization into both $q$ and $r$, while \autoref{alg:ELBO} includes a provision for this case as well.

\begin{algorithm}
\caption{Unbiased $L$-layer CIF ELBO estimator}
\label{alg:ELBO}
\begin{algorithmic}
\FUNCTION{ELBO($x$, amortized)}
    \IF{amortized}
        \STATE $q_0 \gets q_{W_0 | X}(\cdot \mid x)$
    \ELSE
        \STATE $q_0 \gets q_{W_0}$
    \ENDIF
    \STATE $w_0 \sim q_0$
    \STATE $\Delta \gets - \log q_0(w)$
    \FOR{$\ell=1,\ldots,L$}
        \STATE $u \sim q_{U_\ell | W_{\ell-1}}(\cdot \mid w_{\ell-1})$
        \STATE $w_\ell \gets G_\ell(u; w_{\ell-1})$
        \IF{amortized}
            \STATE $r_\ell \gets r_{U_\ell | W_\ell, X}(\cdot \mid w_\ell, x)$
        \ELSE
            \STATE $r_\ell \gets r_{U_\ell | W_\ell}(\cdot \mid w_\ell)$
        \ENDIF
        \STATE $\Delta \gets \Delta + \log r_\ell(u) - \log q_{U_\ell | W_{\ell-1}}(u \mid w_{\ell-1})$ 
        \STATE $\qquad \qquad + \log | \det \mathrm D G_\ell(w_{\ell-1}; u)| $
    \ENDFOR
    \STATE {\bf return} $\Delta + \log p_{X,Z}(x, w_L)$
\ENDFUNCTION
\end{algorithmic}
\end{algorithm}

\section{Comparison to Related Work}

In this section we first compare against methods using explicit normalizing flow models for variational inference, then move on to a discussion of implicit VI methods, and lastly compare the structure of CIFs in basic density estimation to CIFs in VI.

\subsection{Normalizing Flows for VI}

Normalizing flows (NFs) originally became popular as a method for increasing the expressiveness of explicit variational inference models \citep{rezende2015variational}.
NF methods define $q_Z$ as the $Z$-marginal of
\begin{equation} \label{eq:nf_generative}
    W \sim q_W, \qquad  Z = g(W),
\end{equation}
where $g : \Z \rightarrow \Z$ is a bijection.
We can equivalently write $q_Z$ as $q_Z \coloneqq g_\# q_W$, where $g_\# q_W$ denotes the \emph{pushforward} of the distribution $q_W$ under the map $g$.
Using the change of variable formula, we can rewrite \eqref{eq:elbo_1} here as 
\begin{equation} \label{eq:elbo_nf}
    \mathcal L_1(x) = \E_{w \sim q_W} \left[ \log \frac{p_{X, Z}(x, g(w))}{q_W(w) \cdot |\det \mathrm D g(w)|^{-1}}\right].
\end{equation}
This objective is a simplified version of the CIF VI objective \eqref{eq:elbo_cif}.
The following proposition, which we adapt here to the VI setting from \citet[Proposition~4.1]{cornish2019relaxing}, shows that generalizing from \eqref{eq:elbo_nf} to \eqref{eq:elbo_cif} is 
beneficial, as a CIF model trained by this auxiliary bound will perform at least as well in inference as its corresponding baseline flow trained via maximization of \eqref{eq:elbo_nf}.

\begin{proposition} \label{prop:cif_vs_baseline}
Assume a CIF inference model with components $q^\phi_{U|W}$, $r^\phi_{U|Z}$, and $G_\phi$ is parametrized by $\phi \in \Phi$, with associated objective \eqref{eq:elbo_cif} denoted as $\mathcal L_2^\phi$.
Suppose there exists $\psi \in \Phi$ such that for some bijection $g$, $G_\psi(\cdot; u) = g(\cdot)$ for all $u \in \mathcal U$.
Similarly, suppose $q_{U|W}^\psi$ and $r_{U|Z}^\psi$ are such that, for some density $\rho$ on $\mathcal U$, $q_{U|W}^\psi(\cdot \mid w) = r_{U|Z}^\psi(\cdot \mid z) = \rho(\cdot)$ for all $w, z \in \mathcal Z$.
For a given $x \in \X$, if $\mathcal L_2^\phi(x) \geq \mathcal L_2^\psi(x)$,
\[
    \KL{q_Z^\phi}{p_{Z|X}(\cdot \mid x)} \leq \KL{g_\# q_W}{p_{Z|X}(\cdot \mid x)}.
\]
\end{proposition}

The proof of this result, from which we also see that $\mathcal L_2^\psi(x) = \mathcal L_1(x)$ (where $\mathcal{L}_1(x)$ is as written in \eqref{eq:elbo_nf}) for all $x \in \mathcal X$, is provided in \autoref{sec:app-generalization}.
This shows that optimizing a CIF using the auxiliary ELBO $\mathcal L_2$ will produce at least as good of an inference model (as measured by the KL divergence) as a baseline normalizing flow optimized using the marginal ELBO \eqref{eq:elbo_nf}, in the limit of infinite samples from the inference model.
Note that our choices of $G$ from \eqref{eq:cif_G} and $q_{U|W}$ and $r_{U|Z}$ as conditionally Gaussian will usually entail the conditions of \autoref{prop:cif_vs_baseline}, since for example we have $G(w; u) = g(w)$ in \eqref{eq:cif_G} if the final layer weights in the $s$ and $t$ networks are zero.
We also empirically confirm that \autoref{prop:cif_vs_baseline} holds in the experiments.

Beyond the discussion above, we also note that the bijectivity constraint of baseline normalizing flows can lead to problems when modelling a density that is concentrated on a region with complicated topological structure \citep[Corollary~2.2]{cornish2019relaxing}, and may cause flows to become numerically non-invertible in this case \citep{behrmann2020on}.
Many models such as neural spline flows (NSFs) \citep{durkan2019neural} and \emph{universal} flows \citep{huang2018neural,jaini2019sum} have been proposed to improve expressiveness within the standard framework based on a single bijection.
CIFs, on the other hand, use auxiliary variables to provide a mechanism for circumventing the limitations of using a single bijection, but lose analytical tractability as a result.

\subsection{Implicit VI Methods}

Several other AVI methods exist that, like our approach, also require the specification of parametrized auxiliary inference distribution $r_{U|Z}$.
Hierarchical variational models (HVMs) \citep{ranganath2016hierarchical} are one such example, which take $q_{Z,U}(z,u) \coloneqq q_{Z \mid U} (z \mid u) \cdot q_U(u)$ for parametrized distributions $q_{Z \mid U}$ and $q_U$ both analytically tractable.
Although both CIFs and HVMs specify tractable $q_{Z,U}$, the CIF joint distribution \eqref{eq:cif_joint} does not admit such a simple factorization, which may therefore increase expressiveness.
Furthermore, unlike CIFs, HVMs do not admit a natural mechanism for matching the auxiliary inference model $r_{U|Z}$ to the structure of the true auxiliary posterior $q_{U\mid Z}$ when considering multiple levels of hierarchy. %

Related to these are approaches are Hamiltonian-based VI methods \citep{salimans2015markov, caterini2018hamiltonian}, which build $q_{Z,U}$ by numerically integrating Hamiltonian dynamics,
inducing a flow that is bijective now on the extended space $\mathcal Z \times \mathcal U$ instead of just $\Z$.
In contrast, CIFs can be used to augment any type of normalizing flow (not just Hamiltonian dynamics), and are not restricted to a specific family of bijections $G$.
Hamiltonian methods also suffer from greatly increasing computational requirements as the number of parameters in $p_{X,Z}$ grows, since they require $\mathrm D_z \log p_{X,Z}(x,z)$ at every flow step.

There also exist methods which that do not parametrize $r_{U\mid Z}$, but instead build an auxiliary inference distribution in VI by drawing extra samples from the approximate posterior $q_Z$ and re-weighting (as noted in \citet{lawson2019energy}).
These methods, including the importance-weighted autoencoder (IWAE) \citep{DBLP:journals/corr/BurdaGS15} and semi-implicit variational inference \citep{yin2018semi}, effectively perform inference over an extended space consisting of $K$ copies of the original latent space \citep{domke2018importance}.
These approaches may thus require far more memory to train than parametrized AVI methods, and often require care to ensure the variance of estimators of the objective (and gradients thereof) is controlled \citep{rainforth2018tighter, DBLP:conf/iclr/TuckerLGM19}.
That being said, it may be possible to combine multi-sample bounds with CIF models using a framework such as the one in \citet{sobolev2019importance}, which demonstrates how to use IWAE-like approaches within HVMs.

A separate class of implicit VI models proposes expressive
but intractable joint densities requiring density ratio
estimation to train \citep{huszar2017variational, tran2017hierarchical}. CIFs, along with other AVI methods, avoid density ratio estimation by instead constructing a tractable joint density $q_{Z,U}$.

\subsection{CIFs for Density Estimation} \label{sec:cif_for_de}

As mentioned earlier, CIFs were originally proposed as a model for density estimation (DE), a setting in which we have access to a set of observed data $\{x_i\}_i$ over which we would like to build a density model $p_X$ maximizing the marginal likelihood.
This constitutes the key distinction between this work and \citet{cornish2019relaxing}: here, we only use CIFs for parametrizing an \emph{inference} model $q_Z$, assuming we \emph{already} have access to a forward density model $p_{X,Z}$.

However, the inference procedure required to \emph{train} CIFs for DE is actually very closely related to the model \eqref{eq:cif_generative}.
In particular, if we relabel the forward CIF model for DE as $r$ (instead of $p$ used by \citet{cornish2019relaxing}), the single-layer CIF density estimation objective is equivalent to
\begin{equation} \label{eq:cif_de_vs_vi}
    \E_{(x,u) \sim q_{X\!, U}}\!\! \left[ \log\! \frac{r_Z(G(x; u))\cdot r_{U|Z}(u \mid G(x; u))}{q_X^*(x) \!\cdot\!  q_{U|X}(u \mid x)\! \cdot\! |\det \mathrm D_x G(x; u)|^{-\!1\!}} \!\right]\!,
\end{equation}
where $q_X^*$ is the unknown data-generating distribution from which we have i.i.d.\ samples, and $q_{X, U}(x, u) \coloneqq q^*_X(x) \cdot q_{U|X}(u \mid  x)$.
See \autoref{sec:app-cif_de_vi} for a derivation.
Comparing this with \eqref{eq:elbo_cif}, we see that CIFs for density estimation may be interpreted as performing AVI targeting $r_Z$ with an amortized inference model defined as the $Z$-marginal of
\begin{equation} \label{eq:cif_inference_de}
    X \sim q^*_{X}, \quad U \sim q_{U | X}(\cdot \mid X), \quad Z = G(X; U).
\end{equation}

Furthermore, despite the aesthetic similarities between (7) of \citet{cornish2019relaxing} defining $p$ for DE, and \eqref{eq:cif_generative} here defining $q$ for VI, it is actually the $q$ models that share a natural correspondence with each other.
In both cases, $q$ refers to an inference model that must be reparametrized, whereas neither $p$ in DE nor $r$ here require this.
We might even consider using a CIF as the inference distribution for a CIF density model, which may yield additional benefits from added compositionality, although we leave these considerations as future work.

\begin{figure*}
\begin{minipage}{.33\textwidth}
  \begin{subfigure}{\linewidth}
    \centering
    \includegraphics[width=0.9\linewidth]{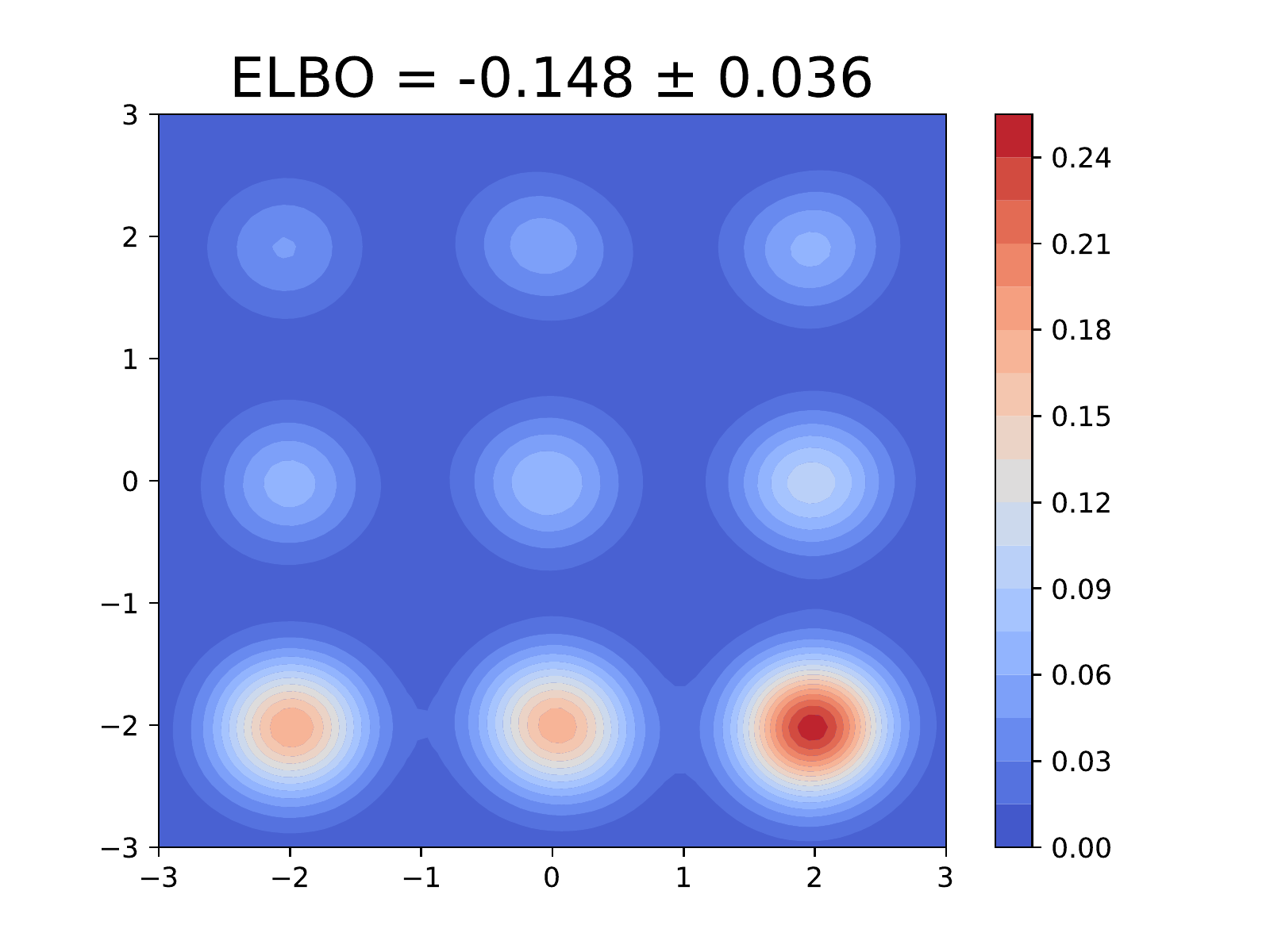}
  \end{subfigure}
  \begin{subfigure}{\linewidth}
    \centering
    \includegraphics[width=0.9\linewidth]{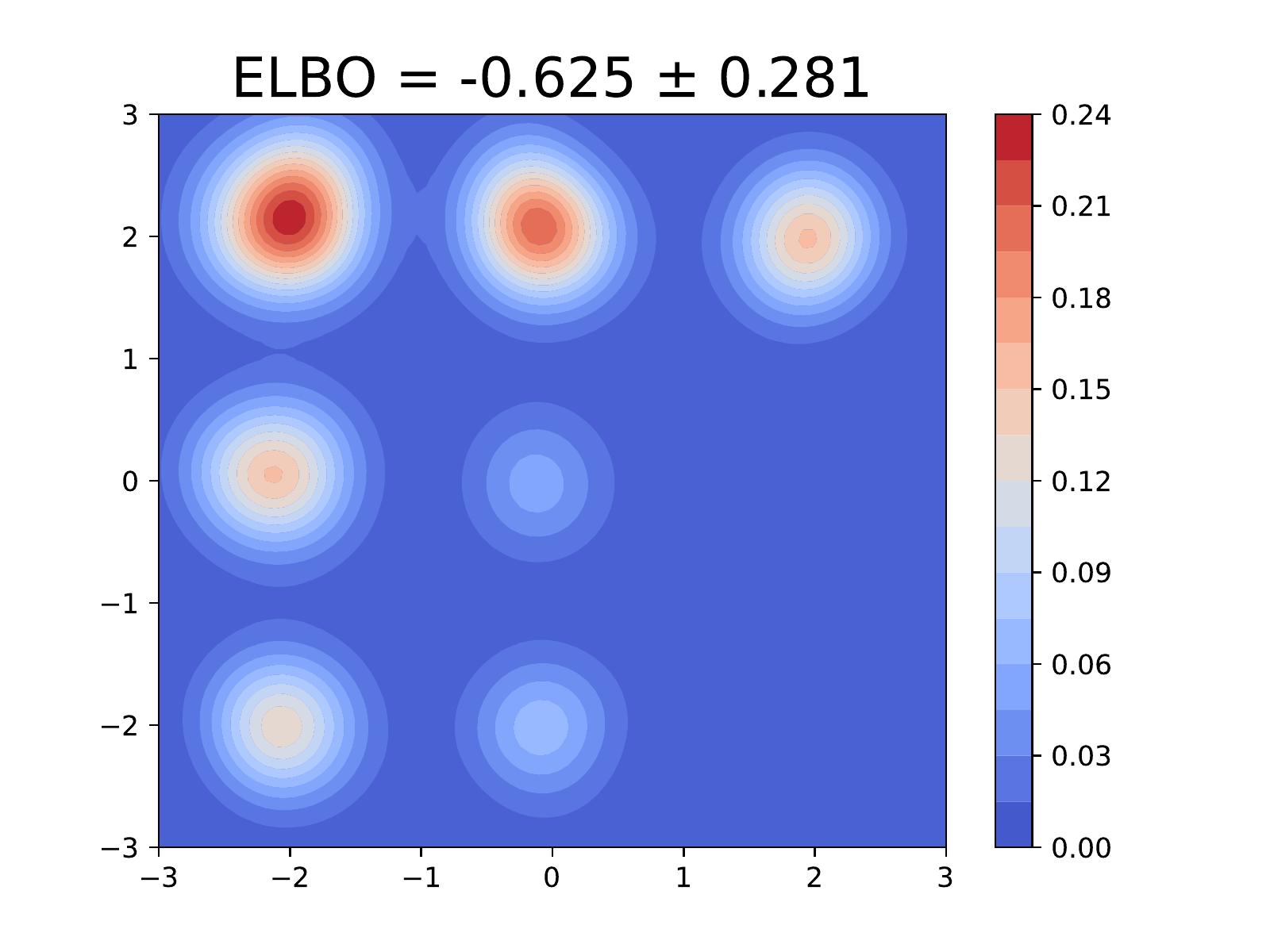}
  \end{subfigure}
\end{minipage}%
\begin{minipage}{.33\textwidth}
  \begin{subfigure}{\linewidth}
    \centering
    \includegraphics[width=0.9\linewidth]{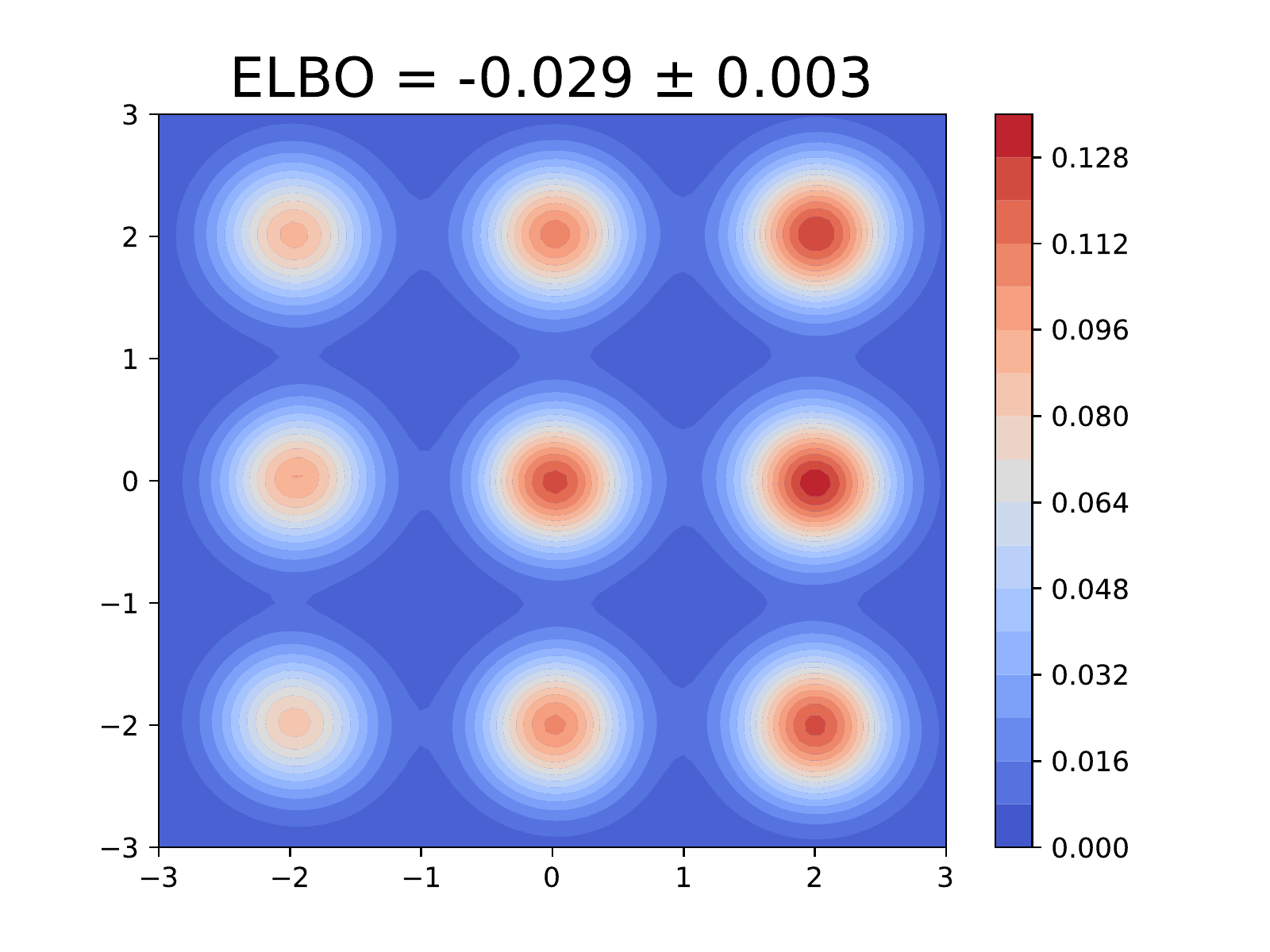}
  \end{subfigure}
  \begin{subfigure}{\linewidth}
    \centering
    \includegraphics[width=0.9\linewidth]{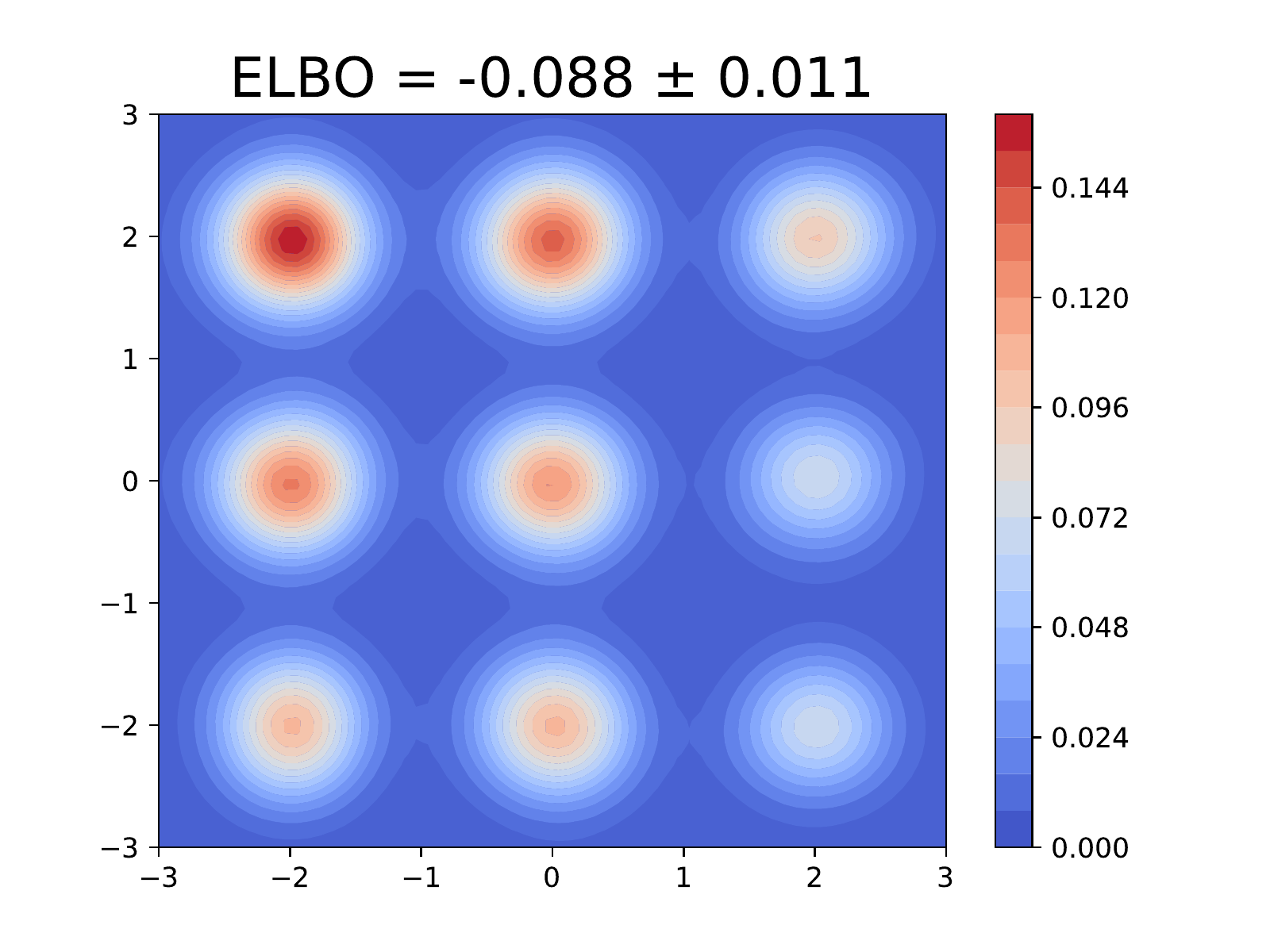}
  \end{subfigure}
\end{minipage}%
\begin{minipage}{.33\textwidth}
  \begin{subfigure}{\linewidth}
    \centering
    \includegraphics[width=0.9\linewidth]{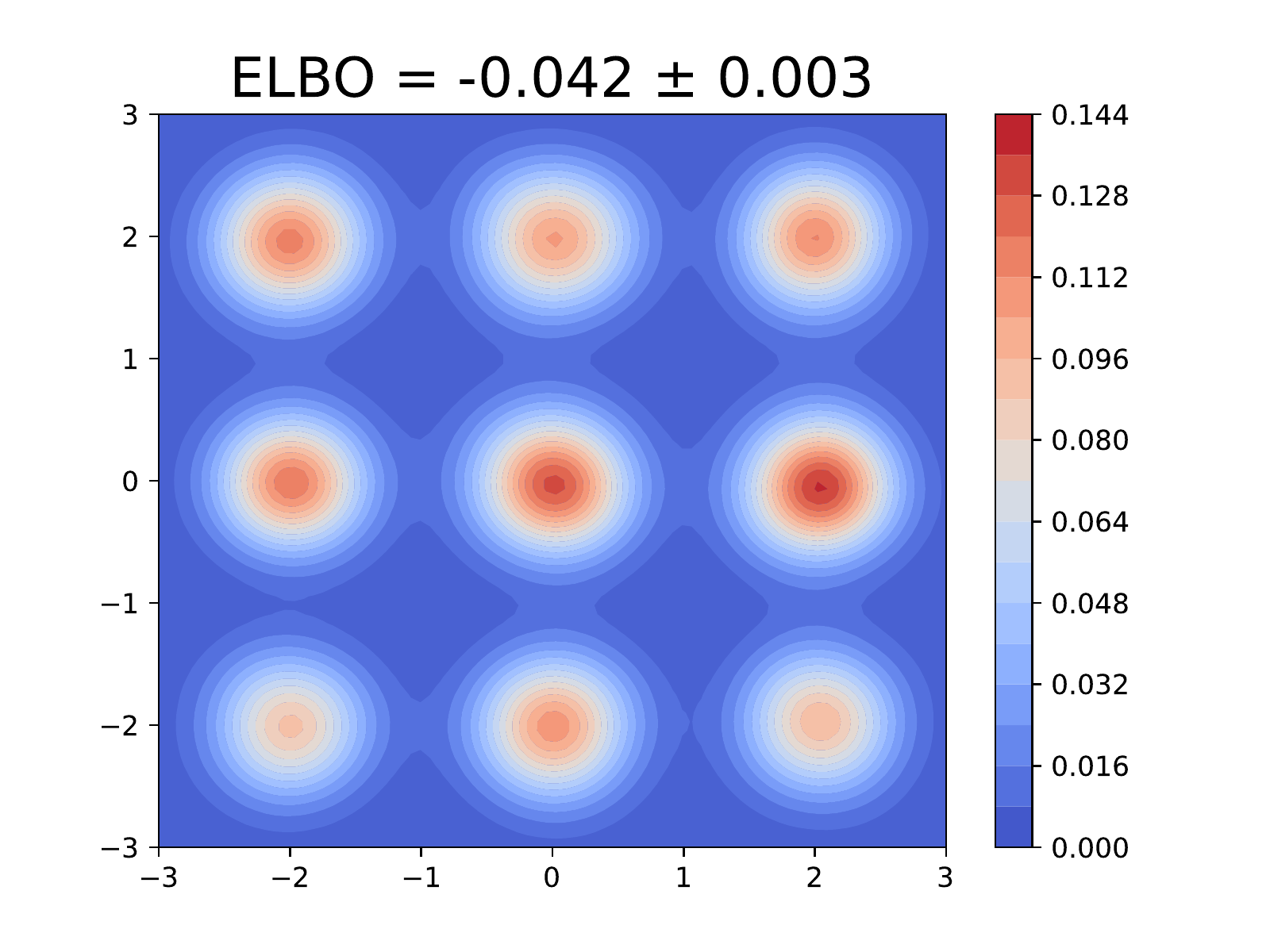}
  \end{subfigure}
  \begin{subfigure}{\linewidth}
    \centering
    \includegraphics[width=0.9\linewidth]{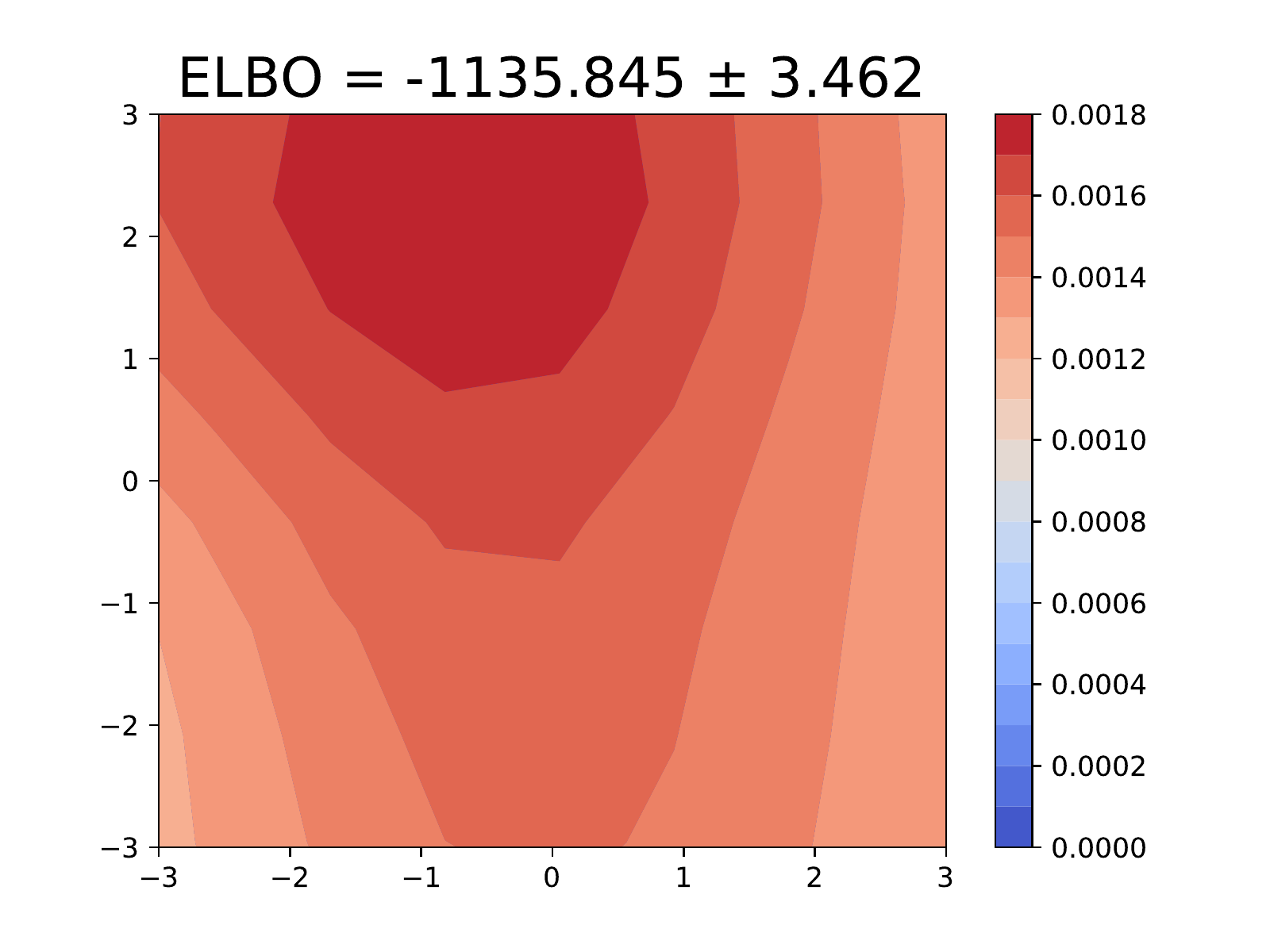}
  \end{subfigure}
\end{minipage}%
\put(-430,-111){$\sigma_0 = 0.1$}
\put(-265,-111){$\sigma_0 = 1$}
\put(-107,-111){$\sigma_0 = 10$}
\vspace{-0.5em}
\caption{
Samples from the trained inference models visualized using a KDE plot for a range of $\sigma_0$ values. 
We ran each configuration 3 times, displaying the average case of the three runs in the image, with the average plus/minus standard error of the marginal ELBO across the three runs shown in the title of the plot (higher is better). 
Models in the top row are CIF-NSFs, and those in the bottom row are baseline NSFs.
We can see that when $\sigma_0 = 0.1$, the NSF does not have enough initial noise to consistently cover the target, and when $\sigma_0 = 10$, the NSF has too much noise and cannot locate the target.
The CIF-NSF at least locates each mode in all cases and provides higher-quality approximations across the board.}
\label{fig:mog-9-components}
\end{figure*}

\section{Experiments} 

In this section, we investigate using CIFs to build more expressive variational models in posterior sampling and maximum likelihood estimation of generative models.
We compare inference models based on the Masked Autoregressive Flow (MAF) \citep{papamakarios2017masked} and the autoregressive variant of the Neural Spline Flow (NSF) \citep{durkan2019neural} to CIF-based extensions.
Both of these baseline models empirically provide good performance in general-purpose density estimation.
We use the ADAM optimizer \citep{DBLP:journals/corr/KingmaB14} throughout.
Hyperparameters for all experiments are available in \autoref{sec:experiment-details}.
Code will be made available at \url{https://github.com/anthonycaterini/cif-vi}. 

\subsection{Toy Mixture of Gaussians} \label{sec:mog}

Our first example looks at using VI to sample from a toy mixture of Gaussians.
Given component means $\{\mu_k\}_k$ and covariances $\{\covmat_k\}_k$, we directly define the ``posterior''\footnote{Note that there is no data $x$ in this example -- we define the ``posterior'' directly. 
Details are in \autoref{sec:experiment-details}.} $p_{Z|X}(z \mid x) \coloneqq \sum_{k=1}^K \mathcal N (z;\mu_k, \covmat_k) / K$,
where $K$ is the total number of components, so that the joint target is $p_{X,Z}(x, z) \propto p_{Z|X}(z \mid x)$.
We work in two dimensions with component means adequately spaced out in a square lattice.
Although the support of $p_{Z|X}$ is all of $\R^2$, it is concentrated on a subset of $K$ disconnected components, which is not homeomorphic to $\R^2$, and thus we anticipate difficulties in using just a normalizing flow as the approximate posterior.
We compare baseline NSF models to CIF-based extensions.

The initial distribution for both the NSF and CIF models is given by $q_W \coloneqq \mathcal N(0, \sigma_0^2 \mathbf I)$, with $\sigma_0$ taken as either a fixed hyperparameter or a trainable variational parameter.
The CIF extension includes an auxiliary variable $u \in \R$ at each layer, conditional Gaussian distributions for $q_{U_\ell|W{\ell-1}}$ and $r_{U_\ell | W_\ell}$ parametrized by small neural networks, and a single small two-headed neural network to output $s$ and $t$ in \eqref{eq:cif_G} at each layer, adding only $8.5\%$ more parameters on top of the baseline NSF model.

\paragraph{Marginal ELBO Estimator} For all experiments in this section, we will measure the trained models on estimates of the marginal ELBO \eqref{eq:elbo_1}.
When using an explicit variational method, such as an NSF, this is readily estimated by basic Monte Carlo (MC) with $N$ i.i.d.\ samples $z^{(i)} \sim q_Z$ for $i \in \{1, \ldots, N\}$:
\begin{equation} \label{eq:explicit_mc_estimator}
    \widehat{\mathcal L}(x) \coloneqq \frac 1 N \sum_{i=1}^N \log \frac{p_{X, Z}(x, z^{(i)})}{q_Z(z^{(i)})}.
\end{equation}
However, recall that in implicit methods $q_Z$ is not available in closed form, which precludes direct evaluation of \eqref{eq:explicit_mc_estimator}.
Thus, we first must build an estimator of $q_Z(z)$ for all $z \in \mathcal Z$ to use within \eqref{eq:explicit_mc_estimator}.
We can do this via importance sampling, taking $M$ i.i.d.\ samples $u^{(j)} \sim r_{U|Z}(\cdot \mid z)$ for $j \in \{1,\ldots, M\}$ from our trained auxiliary inference model:
\begin{equation} \label{eq:q_estimator}
    q_Z(z) \approx \frac 1 M \sum_{j=1}^M \frac{q_{Z,U}(z, u^{(j)})}{r_{U|Z}(u^{(j)} \mid z)} \eqqcolon \widehat q_Z(z).
\end{equation}
The full estimator of the marginal ELBO for auxiliary models is then obtained by substituting \eqref{eq:q_estimator} into \eqref{eq:explicit_mc_estimator}; this is written out in full in \autoref{sec:marginal_elbo_estimator}.
Although this estimator is positively biased (because it includes the negative logarithm of an unbiased MC estimator), it is still \emph{consistent}, and its bias is naturally controlled by the training procedure which encourages $r_{U|Z}$ to match the intractable $q_{U|Z}$.
We can mitigate any further bias by increasing $M$ \citep{rainforth2018nesting}.
A table displaying estimates of the marginal ELBO on a single trained model for various choices of $N$ and $M$ is also available in \autoref{sec:marginal_elbo_estimator}; we choose $N = 10{,}000$ and $M = 100$ based on these results.

\paragraph{Results} For our first experiment, we select $K = 9$ and fix $\sigma_0$ throughout training to either $0.1, 1,$ or $10$.
We train both NSF baselines and CIF-NSF extensions with three different random seeds for each setting of $\sigma_0$.
We show a kernel density estimation of the approximate posterior of the average case model on each configuration in \autoref{fig:mog-9-components} and report the average of the marginal ELBO estimates across all three runs in the titles of the plots.
We can clearly see from both the ELBO values and the plots themselves that the CIF extensions are more consistently producing higher-quality variational approximations across the range of $\sigma_0$, as form of \eqref{eq:cif_G} allows the model to directly control the noise of the outputted samples.
The NSF baselines only produce reliable models for $\sigma_0=1$.

In this example it is quite clear how the parametrization of the CIF model ``cleans up'' a major deficiency of the baseline method by rescaling the initial noise.
However, we might also allow $\sigma_0$ to be learned as part of the overall variational inference procedure to further probe the effectiveness of CIFs, and we experiment with this on a more challenging problem ($K = 16$).
We find that the trained CIF models again outperform the baseline NSFs (estimated marginal ELBO over $3$ runs of $\bf{-0.116 \pm 0.021}$ for CIFs vs.\ $-0.562 \pm 0.008$ for basleine NSFs), thus demonstrating the increased expressiveness of CIFs beyond just rescaling.

\subsection{Generative Modelling of Images} \label{sec:images}

\begin{table*}[!ht]
\caption{Test-set average marginal log-likelihood (plus/minus one standard error) over three runs.
Runs that are within one standard error of the best-performing model are shown in bold.}
\label{tab:images}
\begin{center}
\begin{tabular}{l|l|l|l|l}
\toprule
\multirow{2}{*}{\textbf{Model}} &
  \multicolumn{2}{c|}{\textbf{Small Target}} &
  \multicolumn{2}{c}{\textbf{Large Target}} \\
& MNIST & Fashion-MNIST & MNIST & Fashion-MNIST \\
\midrule
VAE         & $-94.83 \pm 0.05$    & $-238.54 \pm 0.11$    & $-86.27 \pm 0.04$    & $-229.72 \pm 0.03$ \\
IWAE ($K=5$)     & $-93.14 \pm 0.10$ & $-237.03 \pm 0.05$ &  $-84.23 \pm 0.09$ & $-227.80 \pm 0.02$ \\
\midrule
Small MAF        & $-91.98 \pm 0.19$    & $-237.09 \pm 0.15$   & $-83.41 \pm 0.09$   & $ -228.74 \pm 0.24$     \\
Large MAF      & $-92.68 \pm 0.26$    & $-237.57 \pm 0.03$   & $-83.38 \pm 0.12$    & $-228.72 \pm 0.27$   \\
CIF-MAF     & $-90.87 \pm 0.05$    & ${\bf -236.31 \pm 0.14}$   & ${\bf -82.70 \pm 0.12}$    & ${\bf -227.64 \pm 0.05}$ \\
\midrule
Small NSF        & $-91.12 \pm 0.15$    & $-236.65 \pm 0.17$    & $-83.06 \pm 0.05$    & $-228.58 \pm 0.18$   \\
Large NSF      & ${\bf -90.79 \pm 0.02}$    & ${\bf -236.48 \pm 0.13}$    & $-83.12 \pm 0.10$    & $-228.46 \pm 0.07$   \\
CIF-NSF     & ${\bf -90.82 \pm 0.09}$    & ${\bf -236.48 \pm 0.20}$    & $-83.31 \pm 0.17$    & $-228.54 \pm 0.12$ \\
\bottomrule
\end{tabular}
\end{center}
\end{table*}

For our second example, we use amortized variational inference  to facilitate the training of a generative model of image data in the style of the variational auto-encoder (VAE) method \citep{DBLP:journals/corr/KingmaW13}.  
We attempt to build models of the MNIST \citep{lecun1998gradient} and Fashion-MNIST \citep{xiao2017fashion} datasets, which both contain $256$-bit greyscale images of size $28 \times 28$.
We employ dynamic binarization of these greyscale images at each training step.
The likelihood function to describe an image relies on a neural network ``decoder'' $\pi : \Z \rightarrow [0,1]^d$, such that
\[
    Z \sim \mathcal N(0, \mathbf I), \quad X \sim \bigotimes_{j=1}^d \text{Ber}(\cdot \mid \pi_j(Z))
\]
is the generative process for an image $X$.
In our experiments, we consider two different types of decoders: a small convolutional network with only one hidden layer, and a larger convolutional network with several residual blocks as in e.g.\ \citet{durkan2019neural}.
For the experiments with the smaller decoder, we use a $20$-dimensional latent space $\mathcal Z$, and for the larger decoder, we increase to $32$ dimensions.

\paragraph{Inference Methods} We consider several models of inference to aid in surrogate maximum likelihood estimation of the parameters of $\pi$.
First we consider a VAE inference model, where 
\[
    q_{Z|X}(\cdot \mid x) \coloneqq \mathcal N\left(\mu_Z(x), \text{diag } \sigma^2_Z(x)\right)
\]
with an ``encoder'' neural network taking in image data $x$ and outputting both $\mu_Z$ and $\log \sigma_Z$.
The encoder that we use in all experiments is a single-hidden-layer convolutional network which ``matches'' the structure of the small decoder; we keep the encoder small since the VAE here is just a base upon which we build more complicated inference models.
We also consider an importance-weighted version of this VAE model (IWAE) with $K=5$ importance samples \citep{DBLP:journals/corr/BurdaGS15}, which we find roughly matches the computation time per epoch of the flow-based inference methods below.

The first flow-based model that we consider is a $5$-layer masked autoregressive flow (MAF) \citep{papamakarios2017masked}, which is  equivalent to an inverse autoregressive flow (IAF) \citep{kingma2016improved} when removing the hypernetworks producing the flow parameters.
We also run experiments with a $10$-layer neural spline flow (NSF) \citep{durkan2019neural}, for which we clip the norm of the gradients to a maximum of $5$ -- as suggested for tabular density estimation -- for increased stability of training.
Additional hyperparameter settings for each flow are available in \autoref{sec:experiment-details}.
As alluded to previously, for each of the flow-based methods we will use the small VAE encoder as a base distribution $q_{W_0 | X}$ to project the image data into the dimension of the latent space; we do this rather than using a large VAE encoder as the base distribution in the large target experiments (as is typically done) to force the flow models to handle more of the inference.
We also consider two baseline variants for each model, a larger and smaller version, which we control by changing the number of hidden channels in the autoregressive maps.

Finally, we consider amortized CIF-based extensions of the \emph{smaller} variants of the flow models mentioned above, so that in the end our CIF models have approximately the same total number of parameters as the larger baseline flows.
We use a $2$-dimensional $u$ at each flow step.
We include parametrized conditional Gaussian distributions for $q_{U_\ell | W_{\ell-1}}$ and $r_{U_\ell | W_\ell, X}$ at each layer $\ell \in \{1, \ldots, L\}$, with additional care taken in the structure of the $r$ network to combine vector inputs $W_\ell$ with image inputs $X$ -- details are provided in Appendix~\ref{sec:cif-settings}.
We use a single neural network at each layer to parametrize $s_\ell$ and $t_\ell$ appearing in $G_\ell$.

\paragraph{Results} The results of the experiment are available in \autoref{tab:images}.
We use the standard importance-sampling based estimator of the marginal likelihood from \citet[Appendix~E]{rezende2014stochastic} with $1{,}000$ samples, which we find empirically produces low-variance estimates for the small target model\footnote{We expect the same low-variance behaviour to translate to the larger target model, but did not run this for computational reasons.} as noted in Appendix \ref{sec:marg_ll_est}.
We see that, in each experiment, CIF models are either producing the best average performance as measured by test-set estimated average marginal likelihood, or are within error bars of the best.
Importantly, we note that CIFs are outperforming the baseline models which they are built directly on top of across the board: CIF-MAF and CIF-NSF significantly improve upon Small MAF and Small NSF, respectively.
This justifies the claims of \autoref{prop:cif_vs_baseline}, demonstrating that we are not penalized for using the auxiliary objective instead of the standard ELBO.

We also can see that the CIF models produce better results than the IWAE models, which can themselves be seen as a method for auxiliary VI as previously mentioned.
Despite IWAE methods being more parameter-efficient, we found that increasing $K$ for IWAE significantly increased training time per epoch over the CIF models.

\section{Conclusion and Discussion}

In this work, we have presented continuously-indexed flows (CIFs) as a novel parametrization of an approximate posterior for use within variational inference (VI).
We did this by naturally incorporating the CIF model into the framework of AVI.
We have shown that the theoretical and empirical benefits of CIFs over baseline flow models extend to the VI setting, as CIFs outperform baseline flows in both sampling from complicated target distributions and facilitating maximum likelihood estimation of parametrized latent-variable models.
We now add a brief further discussion on CIFs in VI and consider some directions for future work.

\paragraph{Modelling Discrete Distributions} One issue with CIFs for VI (indeed, CIFs more generally) is that they are currently only designed to model continuous distributions, unlike e.g.\ HVMs.
It may be possible however to alleviate this constraint by using discrete flows \citep{NEURIPS2019_9e9a30b7} as a component of the overall CIF model, although it remains to be seen if the theoretical and empirical benefits of CIFs over baseline flows would extend to this case.

\paragraph{CIFs in Other Applications} This work can serve as a template for applying CIFs more generally in applications where NFs have proven effective, such as compression \citep{ho2019compression} and approximate Bayesian computation \citep{papamakarios2019sequential}.
These approaches may require the formulation of appropriate, application-specific surrogate objectives, but the expressiveness gains could overcome the additional costs (as in VI and density estimation) and could therefore be investigated.

\begin{acknowledgements}
Anthony Caterini is a Commonwealth Scholar supported by the U.K.\ Government.
Rob Cornish is supported by the Engineering and Physical Sciences Research Council (EPSRC) through the Bayes4Health programme Grant EP/R018561/1. 
Arnaud Doucet is supported by the EPSRC CoSInES (COmputational Statistical INference for Engineering and Security) grant EP/R034710/1
\end{acknowledgements}

\bibliography{refs}
\bibliographystyle{plainnat}

\onecolumn

\section*{VARIATIONAL INFERENCE WITH CONTINUOUSLY-INDEXED NORMALIZING FLOWS: SUPPLEMENTARY MATERIAL}

\section{Single-Layer Density and Objective Function} \label{sec:app-density-obj}

Here we demonstrate how to obtain the single-layer CIF density and associated objective function required in \autoref{sec:cif-avi}.

\paragraph{Single-layer density} We can derive the joint density $q_{Z,U}$ by first considering the density over $(Z, U, W)$ and integrating out $W$:
\[
    q_{Z,U}(z, u) = \int q_{Z, U, W}(z, u, w) \, \mathrm dw = \int q_W(w) \cdot q_{U|W}(u \mid w) \cdot \delta(z - G(w; u)) \, \mathrm dw.
\]
Now if we perform the change of variable $w = G^{-1}(z'; u)$, we get $\mathrm d w = |\det \mathrm D_z G^{-1}(z'; u) | \, \mathrm d z'$, which then gives
\begin{align*}
    q_{Z,U}(z, u) &= \int q_W(G^{-1}(z'; u)) \cdot q_{U|W}(u \mid G^{-1}(z'; u)) \cdot \delta (z - z') \cdot |\det \mathrm D_z G^{-1}(z'; u) |\, \mathrm d z' \\
    &= q_W(G^{-1}(z; u)) \cdot q_{U|W}(u \mid G^{-1}(z; u)) \cdot |\det \mathrm D_z G^{-1}(z; u) |.
\end{align*}

\paragraph{Single-layer objective} First, we substitute our model $q_{Z,U}$ into \eqref{eq:elbo_2} to obtain
\[
    \mathcal L(x) = \E_{(z, u) \sim q_{Z, U}} \left[\frac{p_{X,Z}(x,z) \cdot r_{U|Z}(u\mid z)}{q_W(G^{-1}(z; u)) \cdot q_{U|W}(u \mid G^{-1}(z; u)) \cdot |\det \mathrm D_z G^{-1}(z; u) |} \right].
\]
Now, noting that $z = G(w; u)$ for some $w \sim q_W$ as per the sampling procedure \eqref{eq:cif_generative}, we can rewrite the above objective instead as an expectation over $q_{W,U}$ (using LOTUS) to obtain
\[
    \mathcal L (x) = \E_{(w, u) \sim q_{W ,U}} \left[ \log \frac{p_{X, Z}(x, G(w; u)) \cdot r_{U|Z}(u \mid G(w; u))}{q_W (w) \cdot q_{U | W}(u \mid w) \cdot |\det \mathrm D_w G(w; u) |^{-1}}\right],
\]
since $\mathrm D_z G^{-1}\left(G(w; u); u\right) = \mathrm D_w G(w; u)$, which recovers \eqref{eq:elbo_cif}.

\section{Multi-layer Density and Objective Function} \label{sec:app-stacking}

This section is much like the previous section, except this time for the multi-layer model.
We demonstrate how to recursively calculate the density and provide the objective function for a multi-layer model in both the un-amortized and amortized settings.

\paragraph{Recursive multi-layer density} We can derive the full joint density $q_{Z, U_{1:L}}$ by first considering an intermediate density $q_{W_\ell, U_{1:\ell}}$ for $\ell \in \{1, \ldots, L\}$, then integrating over the variable $W_{\ell-1}$:
\begin{align*}
    q_{W_\ell, U_{1:\ell}} (w_\ell, u_{1:\ell}) &= \int q_{W_\ell, U_\ell, W_{\ell-1}, U_{1:\ell-1}}(w_\ell, u_\ell, w_{\ell-1}, u_{1:\ell-1}) \, \mathrm d w_{\ell-1} \\
    &= \int q_{W_{\ell-1}, U_{1:\ell-1}}(w_{\ell-1}, u_{1:\ell-1}) \cdot q_{U_\ell | W_{\ell-1}}(u_\ell \mid w_{\ell-1}) \cdot \delta (w_\ell - G_\ell(w_{\ell-1}; u_\ell)) \, \mathrm d w_{\ell-1},
\end{align*}
where in the second line we use the fact that $U_\ell$ is conditionally independent of $U_{1:\ell-1}$ given $W_{\ell-1}$ (note that this fact is also used to derive the structure of the auxiliary posterior $q_{U_{1:L} | Z}$), and the base case of the recursion is given by $q_{W_0, U_{1:0}} (w_0, -) \coloneqq q_{W_0}(w_0)$.
Now, as in the previous section, we use the change of variable $w_{\ell-1} = G^{-1}_\ell(w_\ell'; u_\ell)$ to obtain $\mathrm d w_{\ell-1} = |\det \mathrm D_{w_\ell} G_\ell^{-1}(w_{\ell}'; u_\ell) | \, \mathrm d w_\ell'$, and thus
\[
    q_{W_\ell, U_{1:\ell}}(w_\ell, u_{1:\ell}) = q_{W_{\ell-1}, U_{1:\ell-1}}\left(G_\ell^{-1}(w_\ell; u_\ell), u_{1:\ell-1} \right) \cdot q_{U_\ell | W_{\ell-1}}\left( u_\ell | G^{-1}_\ell(w_\ell; u_\ell)\right) \cdot |\det \mathrm D_{w_\ell} G_\ell^{-1}(w_\ell; u_\ell)|.
\]
We obtain our full inference model as the $L^{th}$ step of the recursion, i.e.\ $q_{Z, U_{1:L}} \equiv q_{W_L, U_{1:L}}$.

\paragraph{Multi-layer objective function} Given our joint model $q_{Z, U_{1:L}}$ and the factorized auxiliary inference model $r_{U_{1:L} | Z}$ from \eqref{eq:multilayer_backward}, we can write the objective function from \eqref{eq:elbo_2} as
\[
    \mathcal L(x) = \mathbb E_{(z, u_{1:L}) \sim q_{Z, U_{1:L}}} \left[ \log \frac{p_{X,Z}(x,z) \cdot r_{U_{1:L} | Z} (u_{1:L} \mid z)}{q_{Z,U_{1:L}}(z, u_{1:L})} \right].
\]
However, as in the single-layer case, it is difficult to calculate unbiased gradients of the objective -- as written in this form -- with respect to the parameters of the bijections $G_\ell$, as these bijections are also appearing in the distribution over which we take the expectation.
Thus we write $w_\ell = G_\ell(w_{\ell-1}; u_\ell)$ recursively for $\ell \in \{1, \ldots, L\}$, with $z \coloneqq w_L$, to rewrite the objective function instead as an expectation over $q_{W_0, U_{1:L}}(u_{1:L}, w_0) \coloneqq q_{W_0}(w_0) \cdot \prod_{\ell=1}^L q_{U_{\ell} \mid W_{\ell-1}}(u_\ell | w_{\ell-1})$ as below:
\begin{align} \label{eq:cif_multilayer_obj}
    \mathcal L (x) &= \E_{(w_0, u_{1:L}) \sim q_{W_0, U_{1:L}}} \left[\log \frac{p_{X,Z}(x, z) \cdot \prod_{\ell=1}^L r_{U_\ell | W_\ell}(u_\ell \mid w_\ell)}{ q_{W_0}(w_0) \cdot \prod_{\ell=1}^L \left\{ q_{U_\ell | W_{\ell-1}}(u_\ell \mid  w_{\ell-1}) \cdot |\det \mathrm D_{w_{\ell-1}} G_\ell(w_{\ell-1}; u_\ell)|^{-1} \right\} } \right] \nonumber \\
    &= \E_{(w_0, u_{1:L}) \sim q_{W_0,U_{1:L}}} \left[- \log q_{W_0}(w_0) + \sum_{\ell=1}^L \log \frac{r_{U_\ell | W_\ell}(u_\ell \mid w_\ell) \cdot |\det \mathrm D_{w_{\ell-1}} G_\ell(w_{\ell-1}; u_\ell)|}{q_{U_\ell | W_{\ell-1}}(u_\ell \mid  w_{\ell-1})} + \log p_{X,Z}(x, z) \right].
\end{align}
The form of objective given in \eqref{eq:cif_multilayer_obj} demonstrates how \autoref{alg:ELBO} works: initialize with $-\log q_{W_0}(w_0),$ collect $r_\ell, \mathrm D G_\ell, $ and $q_\ell$ terms at each step $\ell$, and then finish by evaluating the joint target at the realized $z$ value. 

\paragraph{Amortization} When using amortization, we can redefine the generative process \eqref{eq:cif_generative_multilayer} given $X$ as follows:
\[
    W_0 \sim q_{W_0 | X}(\cdot \mid X), \qquad U_\ell \sim q_{U_\ell \mid W_{\ell-1}}(\cdot \mid W_{\ell-1}), \qquad W_\ell = G_\ell(W_{\ell-1}; U_\ell),
\]
where $Z \coloneqq W_L$.
Now, additionally conditioning our auxiliary inference model $r_{U_{1:L} | Z, X}$ on $X$, we can write the objective \eqref{eq:cif_multilayer_obj} for the amortized case as 
\begin{equation} \label{eq:a_cif_multilayer_obj}
    \mathcal L(x) = \E_{(w_0, u_{1:L}) \sim q_{W_0, U_{1:L} \mid X}} \left[- \log q_{W_0 | X}(w_0 \mid x) + \sum_{\ell=1}^L \log \frac{r_{U_\ell | W_\ell, X}(u_\ell \mid w_\ell, x) \cdot |\det \mathrm D_{w_{\ell-1}} G_\ell(w_{\ell-1}; u_\ell)|}{q_{U_\ell | W_{\ell-1}}(u_\ell \mid  w_{\ell-1})} + \log p_{X,Z}(x, z) \right].
\end{equation}

\paragraph{Multi-layer CIF as a single-layer CIF}
Lastly, we show here how the multi-layer model \eqref{eq:cif_generative_multilayer} corresponds to an instance of \eqref{eq:cif_generative} for an $L$-layered extended space and bijection (as per \citet[Section~3.1]{cornish2019relaxing}): first define $G^\ell(\cdot; u_1, \ldots, u_\ell) \coloneqq G_\ell(\cdot; u_\ell) \circ \cdots \circ G_1(\cdot; u_1)$, and then take $W \coloneqq W_0, U \coloneqq U_1, \ldots, U_L), q_{U|W}(u \mid w) \coloneqq \prod_\ell q_{U_\ell | W_{\ell-1}}\left(u_\ell \mid G^{\ell}(w; u_{1:\ell})\right),$ and $G \coloneqq G^L$ in \eqref{eq:cif_generative} to arrive at \eqref{eq:cif_generative_multilayer}.

\section{Continuously-Indexed Flows Versus Baseline Normalizing Flows} \label{sec:app-generalization}

Here, we provide a proof of \autoref{prop:cif_vs_baseline}.
Throughout the proof, we consider $x$ such that $\mathcal L_2^\phi(x) \geq \mathcal L_2^\psi(x)$ as per our assumption.

Let us first denote the normalizing flow objective \eqref{eq:elbo_nf} as $\mathcal L_1$.
It is not hard to show that $\mathcal L_2^\psi$ reduces to $\ELBONF$:
\begin{align*}
    \ELBOCIF^\psi(x) &= \E_{(w, u) \sim q^\phi_{W,U}} \left[ \log \frac{p_{X, Z}(x, G_\psi(w; u)) \cdot r_{U|Z}(u \mid G_\psi(w; u))}{q_W (w) \cdot q^\psi_{U | W}(u \mid w) \cdot |\det \mathrm D_w G_\psi(w; u) |^{-1}}\right], \\
    &= \E_{w \sim q_W} \E_{u \sim \rho} \left[ \log \frac{p_{X, Z}(x, g(w)) \cdot \rho(u)}{q_W (w) \cdot \rho(u) \cdot |\det \mathrm D g(w) |^{-1}}\right] \\
    &= \ELBONF(x).
\end{align*}
From \eqref{eq:elbo_difference}, we also know that
\[
    \E_{z \sim q_Z^\phi} \left[ \log p_{X, Z}(x,z) - \log q_Z^\phi(z) \right] \geq \ELBOCIF^\phi(x),
\]
as the left-hand-side is the intractable marginal ELBO obtained when using $q_Z^\phi(z) \coloneqq \int q_{Z,U}^\phi(z,u) \, \mathrm d u$ as the variational distribution.

We get the final result by exploiting the standard relationship between the ELBO and the KL divergence:
\begin{align*}
    \KL{q_Z^\phi}{p_{Z|X}(\cdot | x)} &=  \log p_X(x) - \E_{z \sim q_Z^\phi} \left[ \log p_{X, Z}(x,z) - \log q_Z^\phi(z) \right] \\
    &\leq \log p_X(x) - \ELBOCIF^\phi(x) \\
    &\leq \log p_X(x) - \ELBONF(x) \\
    &= \KL{g_\# q_W}{p_{Z|X}(\cdot | x)}.
\end{align*}

\section{Relationship between Continuously-Indexed Flows for Density Estimation and Variational Inference} \label{sec:app-cif_de_vi}

When we are using CIFs for density estimation, we can write the single-layer generative process as
\[
    Z \sim r_Z, \qquad U \mid Z \sim r_{U|Z}(\cdot \mid Z), \qquad X = G^{-1}(Z; U),
\]
so that $r_X(x) = \int r_{X, U}(x, u)\, \mathrm d u$ is the proposed density model of a dataset $\mathcal D = \{x_i\}_{i=1}^N$, with 
\[
    r_{X,U}(x,u) = r_Z(G(x; u)) \cdot r_{U|Z}(u \mid G(x; u)) \cdot |\det \mathrm D G(x; u)|.
\]
Our goal is to maximize the average likelihood of $r_X(x)$ over the dataset, i.e. $\max \frac{1}{N} \sum_{i=1}^N \log r_X(x_i)$.
However, since $r_X$ is intractable, we must introduce a reparametrizable inference distribution $q_{U|X}$ and instead maximize the ELBO, given here for a datapoint $x \in \mathcal D$:
\begin{equation} \label{eq:elbo_cif_de}
    \mathcal L(x) = \E_{u \sim q_{U|X}(\cdot \mid x)} \left[ \log r_{X, U}(x,u) - \log q_{U|X}(u \mid x) \right].
\end{equation}

Note that instead of maximizing the average of \eqref{eq:elbo_cif_de} over the dataset, we could instead theoretically maximize the average of \eqref{eq:elbo_cif_de} over the unknown ``true'' data-generating distribution -- here denoted $q_X^*$ -- which admits the objective $\max \E_{q_X^*} \mathcal L(x)$.
Maximizing this objective is equivalent to maximizing
\begin{equation} \label{eq:theoretical_cif_de_obj}
    \E_{x \sim q_X^*} \left[\mathcal L(x)\right] - \E_{x \sim q_X^*} \left[\log q_X^*(x)\right]
\end{equation}
since $q_X^*$ is independent of the parameters of the model.
If we substitute \eqref{eq:elbo_cif_de} into this expression and expand the definition for $r_{X,U}$, we have 
\begin{align*}
    &\qquad \E_{x \sim q_X^*} \left[\mathcal L(x)\right] - \E_{x \sim q_X^*} \left[\log q_X^*(x)\right] \\
    &= \E_{x \sim q_X^*} \left[ \E_{u \sim q_{U|X}(\cdot \mid x)} \left[ \log r_{X,U}(x,u) - \log q_{U|X}(u\mid x) \right] - \log q_X^*(x)\right] \\
    &= \E_{x \sim q_X^*} \left[ \E_{u \sim q_{U|X}(\cdot \mid x)} \left[ \log r_Z(G(x; u)) + \log r_{U|Z}(u \mid G(x; u)) + \log |\det \mathrm D_x G(x; u)| - \log q_{U|X}(u \mid x) \right] - \log q_X^*(x)\right]\\
    &= \E_{(x,u) \sim q_{X, U}} \left[ \log \frac{r_Z(G(x; u))\cdot r_{U|Z}(u \mid G(x; u))}{q_X^*(x) \cdot q_{U|X}(u \mid x) \cdot |\det \mathrm D_x G(x; u)|^{-1}} \right],
\end{align*}
which derives \eqref{eq:cif_de_vs_vi}, where we define $q_{X, U}(x, u) \coloneqq q_X^*(x) \cdot q_{U|X}(u\mid x)$.

Note also that maximizing \eqref{eq:theoretical_cif_de_obj} is equivalent to minimizing an upper bound on $\KL{q_X^*}{r_X}$:
\begin{align*}
    \E_{x \sim q_X^*} \left[\log q_X^*(x)\right] - \E_{x \sim q_X^*} \left[\mathcal L(x)\right] &= \E_{x \sim q_X^*} \left[ \log q_X^*(x) -  \E_{u \sim q_{U|X}(\cdot \mid x)} \left[ \log r_{X,U}(x,u) - \log q_{U|X}(u\mid x) \right] \right] \\
    &\geq \E_{x \sim q_X^*} \left[\log q_X^*(x) - \log r_X(x)\right] \quad \text{(Jensen)}\\
    &= \KL{q_X^*}{r_X}.
\end{align*}
This is not surprising but at least motivates the use of \eqref{eq:theoretical_cif_de_obj} as a theoretical objective.

\section{Further Experiment Details} \label{sec:experiment-details}

We have included all details about the experiments from the main text in this section.
We first include the approximate posterior plots from \emph{all} runs of the $K = 9$ mixture of Gaussians experiment and not just the best of three random seeds. 
We next discuss the setup of both the image and mixture of Gaussians problems, then the specific structures used to build the baseline flow models, CIF extensions, and VAE models, then discuss the details of the optimization, and finally describe the log-likelihood estimator used to generate the values in \autoref{tab:images}.

\subsection{More 2D Mixture Plots} \label{sec:app-2d-plots}

\begin{figure*}
\begin{minipage}{.33\textwidth}
  \begin{subfigure}{\linewidth}
    \centering
    \includegraphics[width=0.9\linewidth]{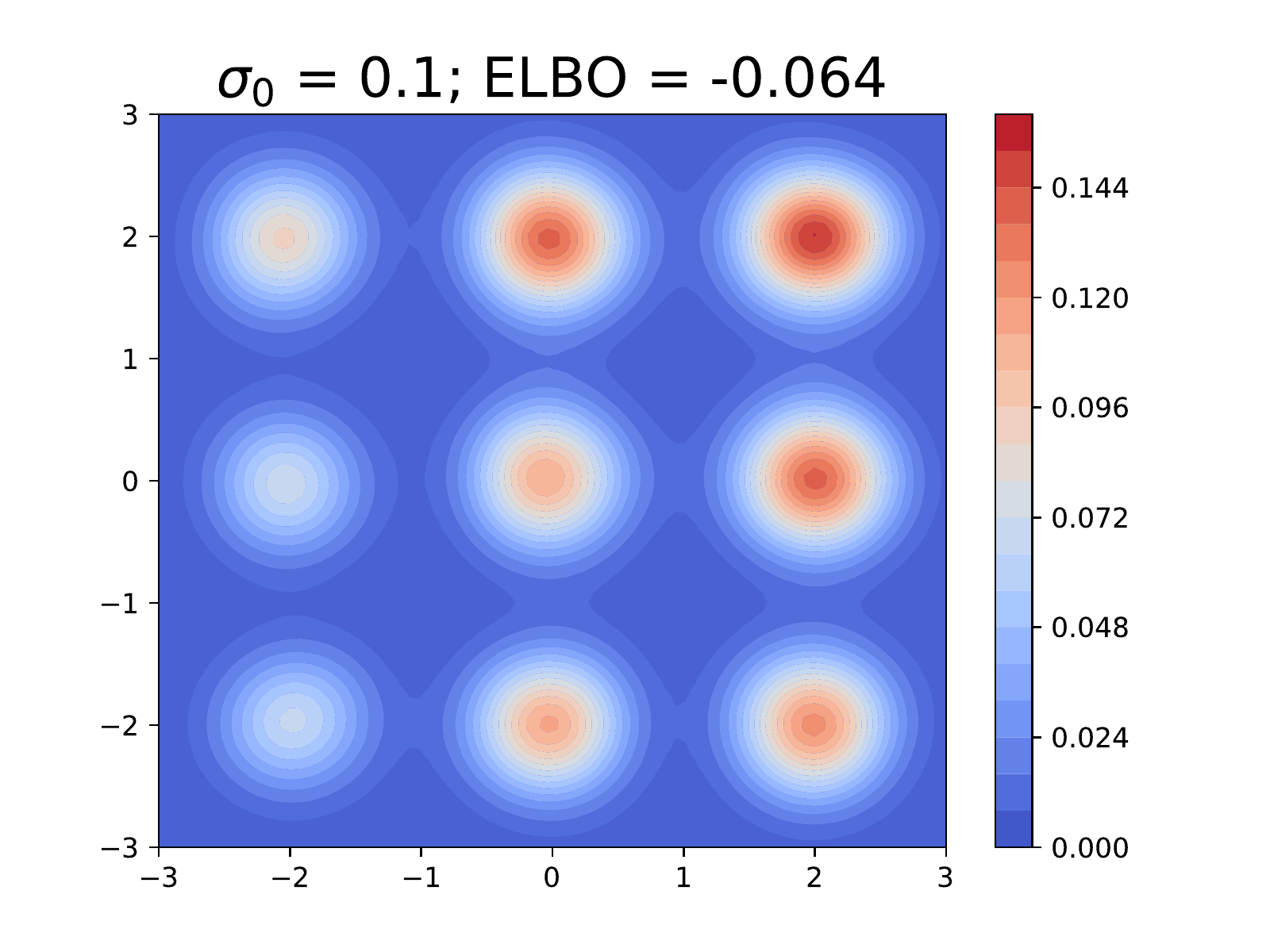}
  \end{subfigure}
  \begin{subfigure}{\linewidth}
    \centering
    \includegraphics[width=0.9\linewidth]{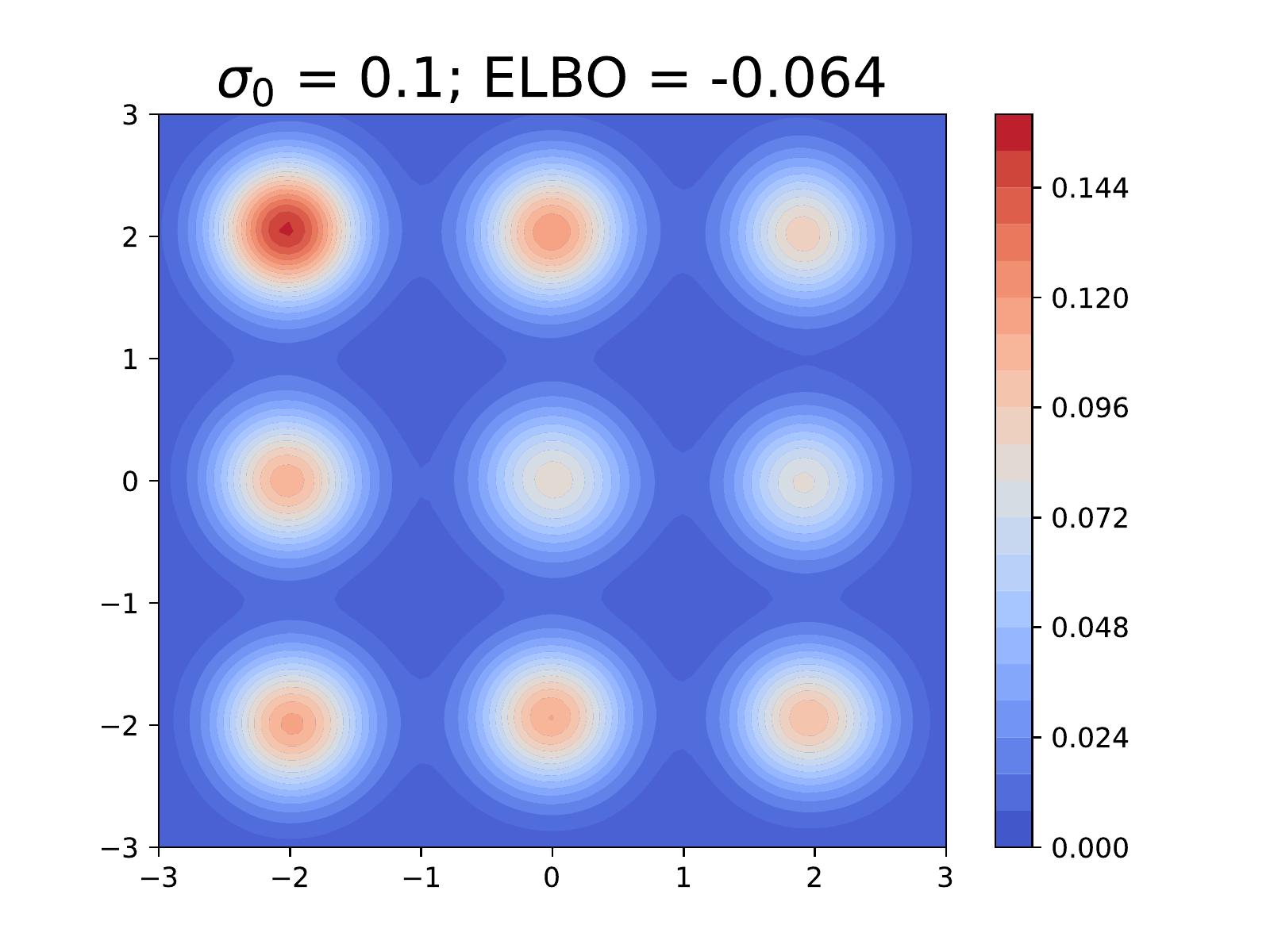}
  \end{subfigure}
\end{minipage}%
\begin{minipage}{.33\textwidth}
  \begin{subfigure}{\linewidth}
    \centering
    \includegraphics[width=0.9\linewidth]{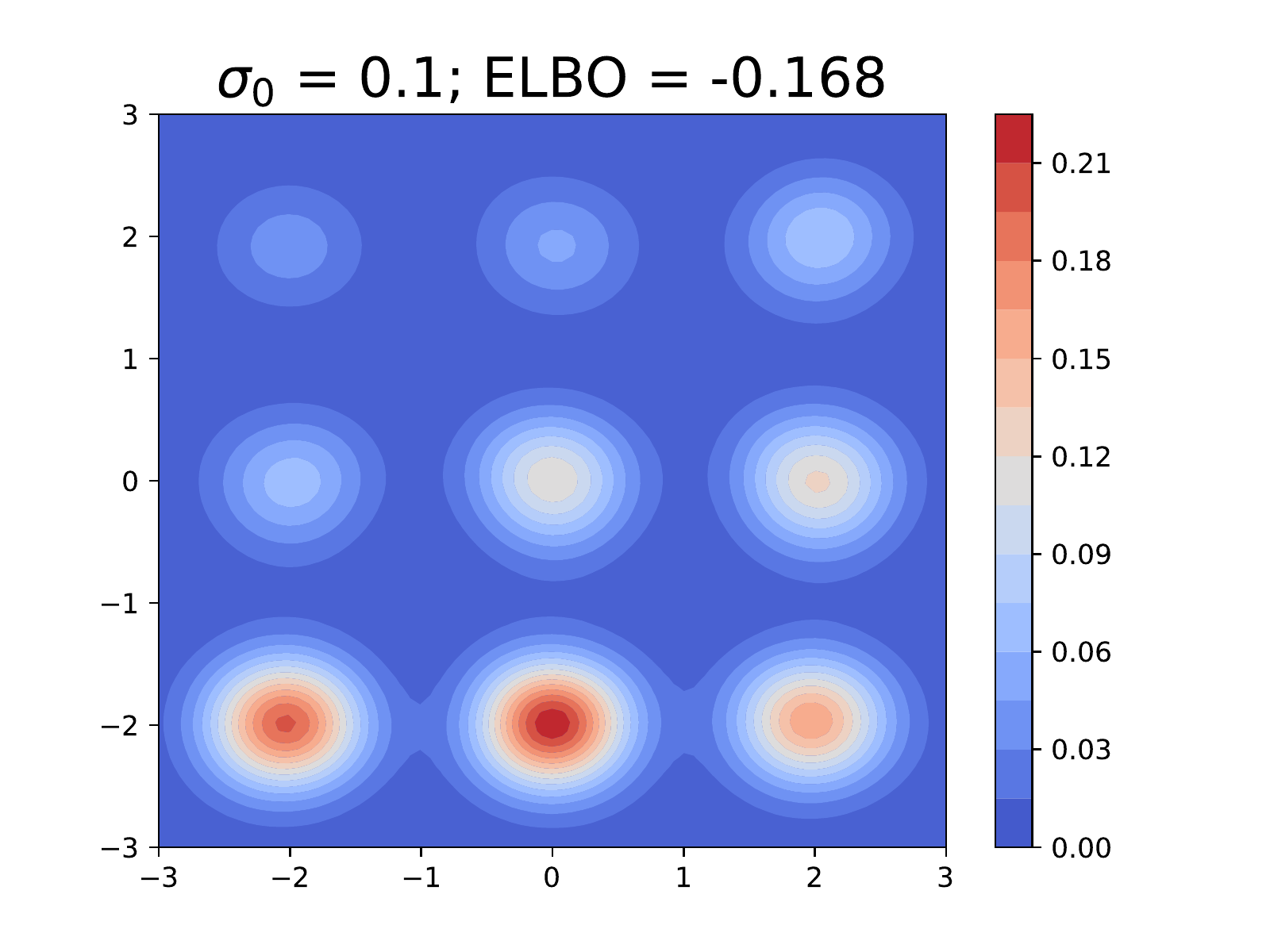}
  \end{subfigure}
  \begin{subfigure}{\linewidth}
    \centering
    \includegraphics[width=0.9\linewidth]{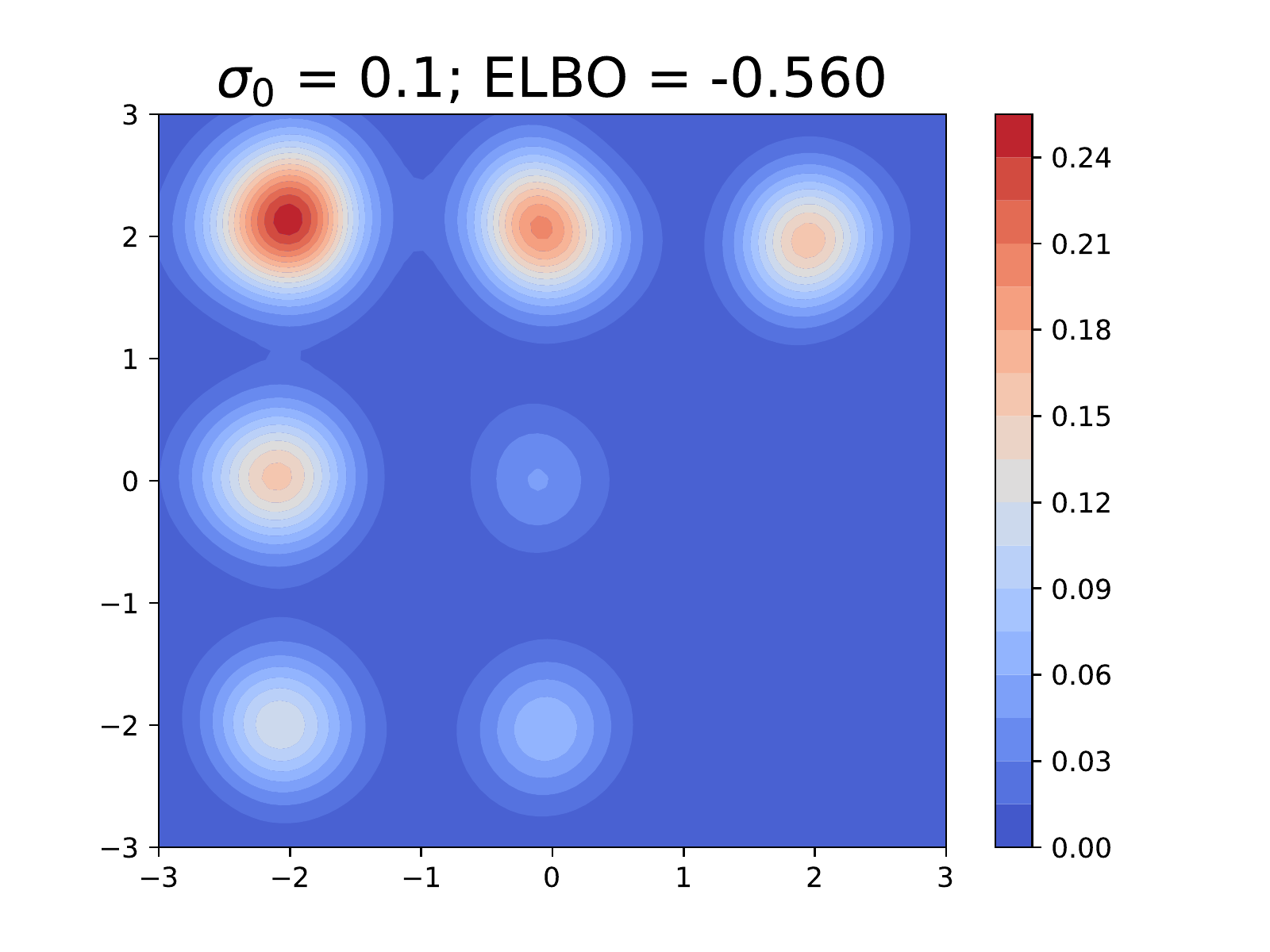}
  \end{subfigure}
\end{minipage}%
\begin{minipage}{.33\textwidth}
  \begin{subfigure}{\linewidth}
    \centering
    \includegraphics[width=0.9\linewidth]{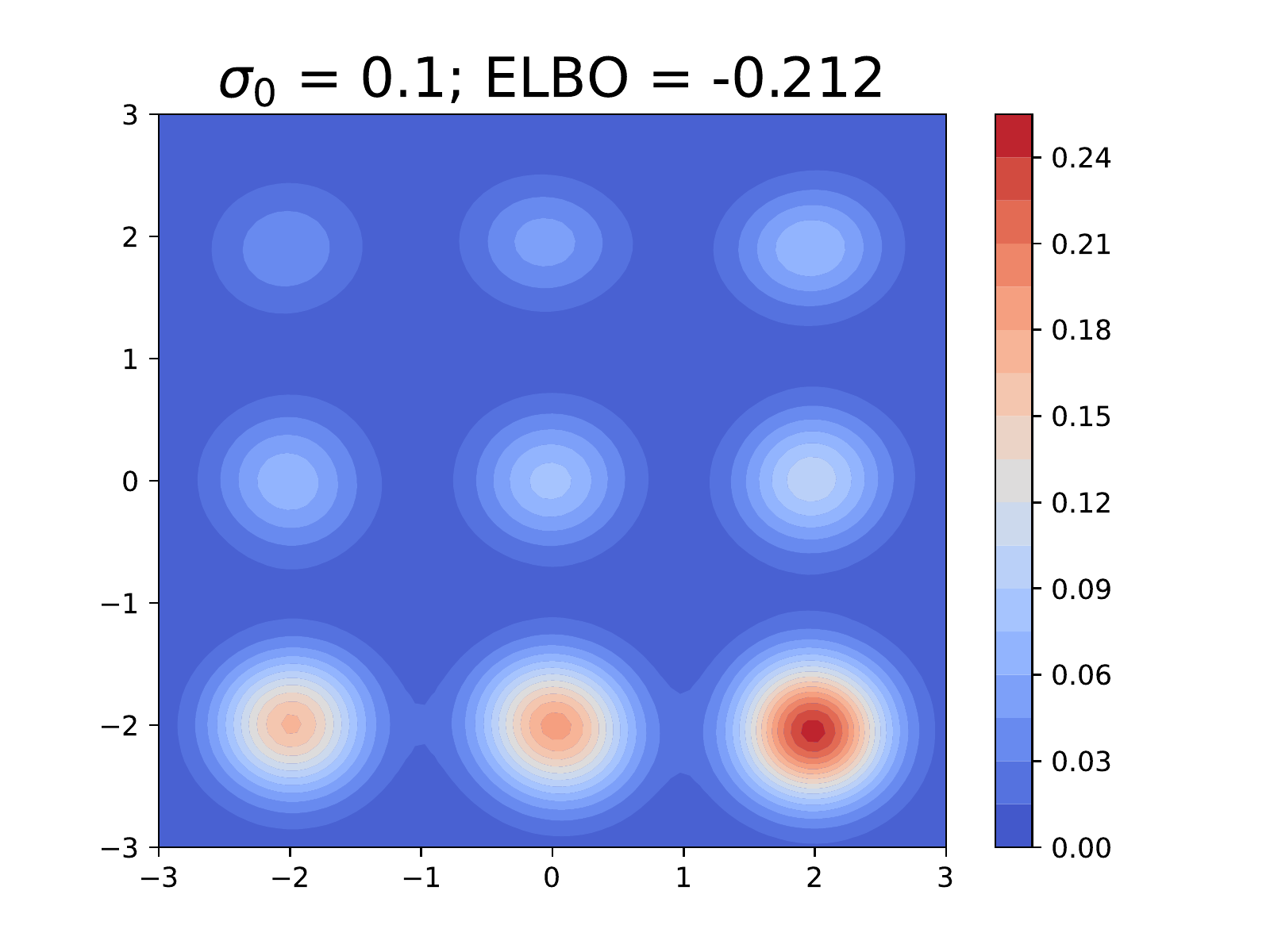}
  \end{subfigure}
  \begin{subfigure}{\linewidth}
    \centering
    \includegraphics[width=0.9\linewidth]{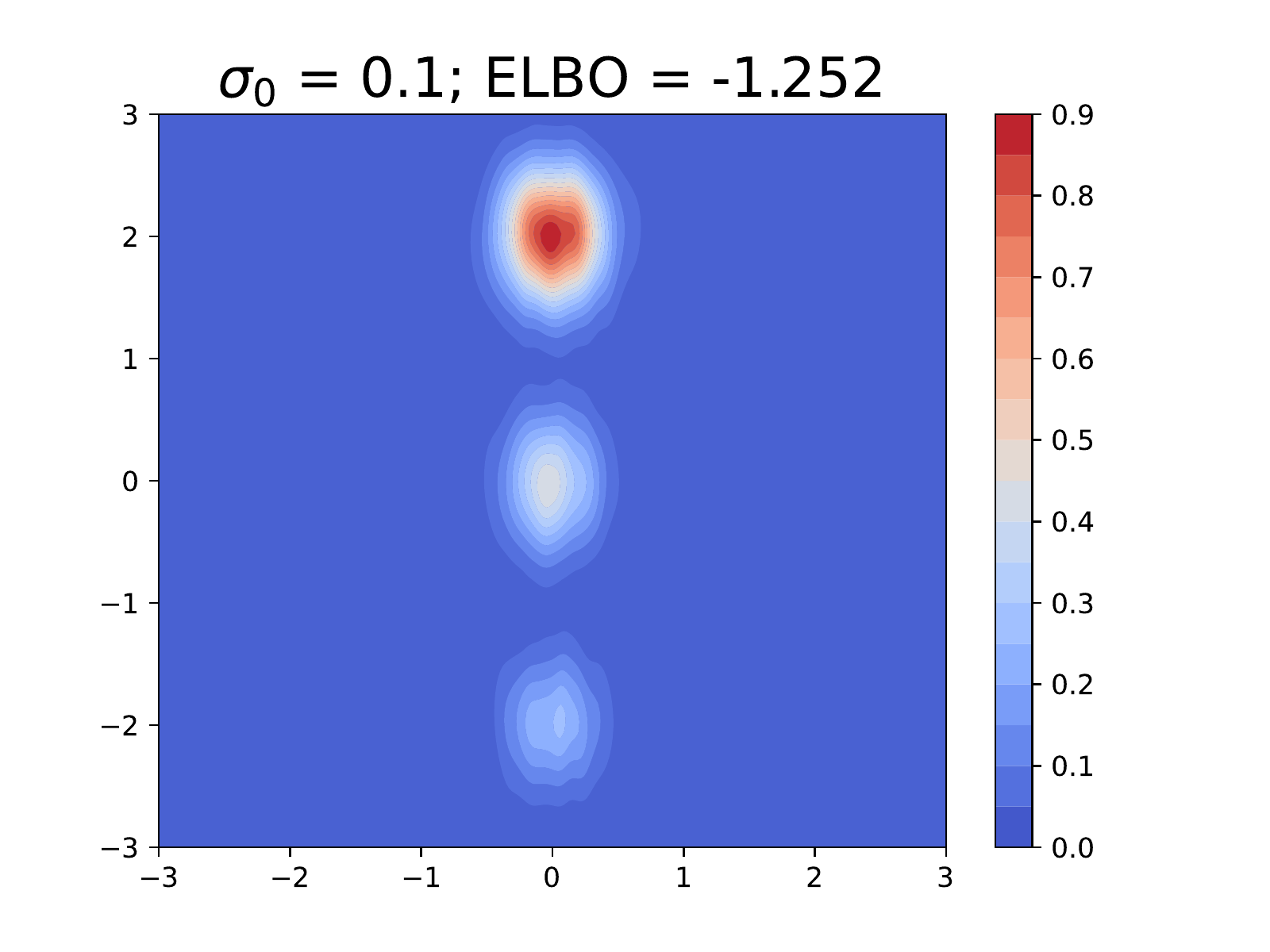}
  \end{subfigure}
\end{minipage}%
\caption{All runs of the mixture of Gaussians experiment for $\sigma_0 = 0.1$.
CIF models are in the top row, NSF models in the bottom row.}
\label{fig:all-mog-small-sigma0}
\end{figure*}

\begin{figure*}
\begin{minipage}{.33\textwidth}
  \begin{subfigure}{\linewidth}
    \centering
    \includegraphics[width=0.9\linewidth]{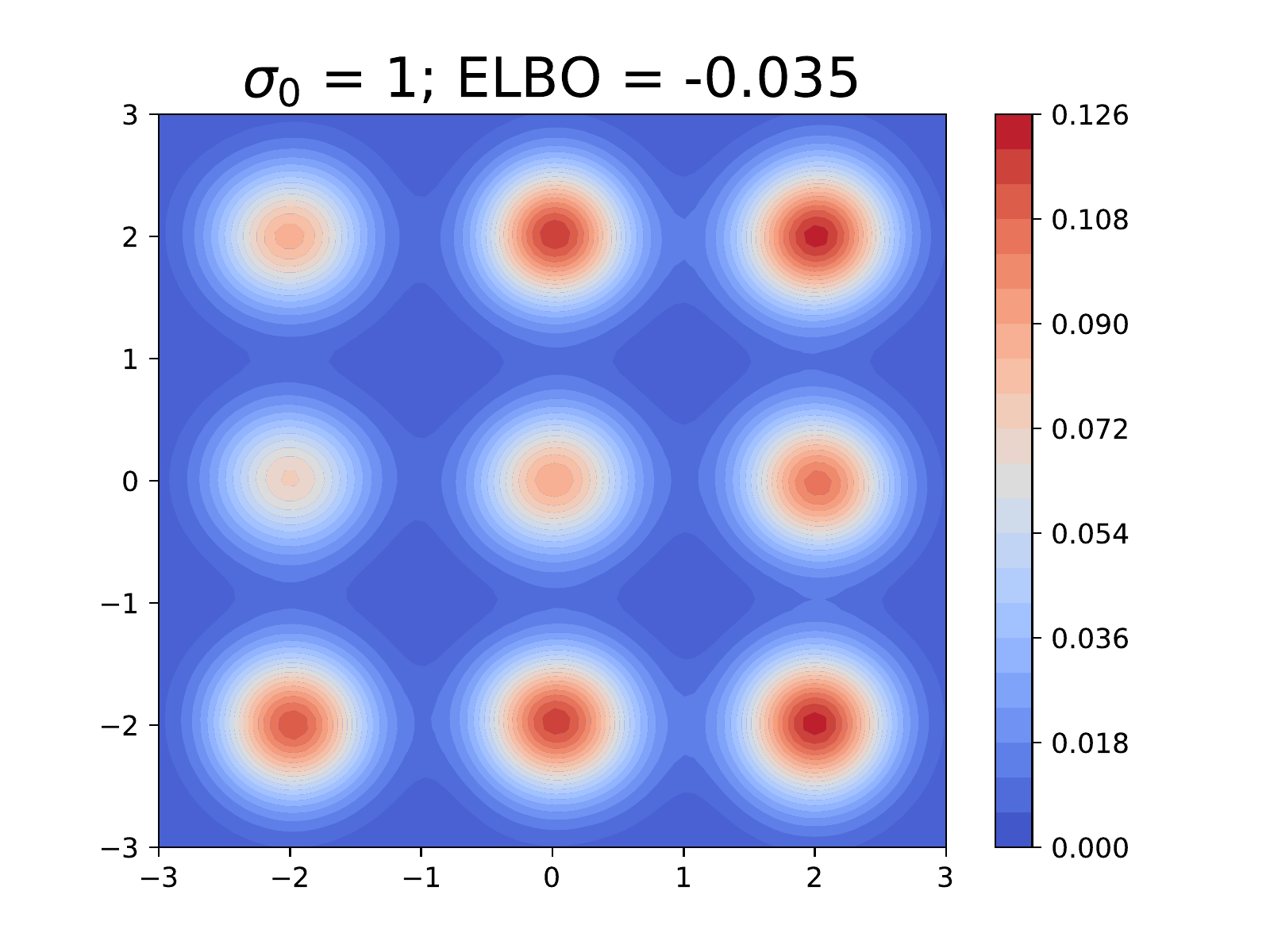}
  \end{subfigure}
  \begin{subfigure}{\linewidth}
    \centering
    \includegraphics[width=0.9\linewidth]{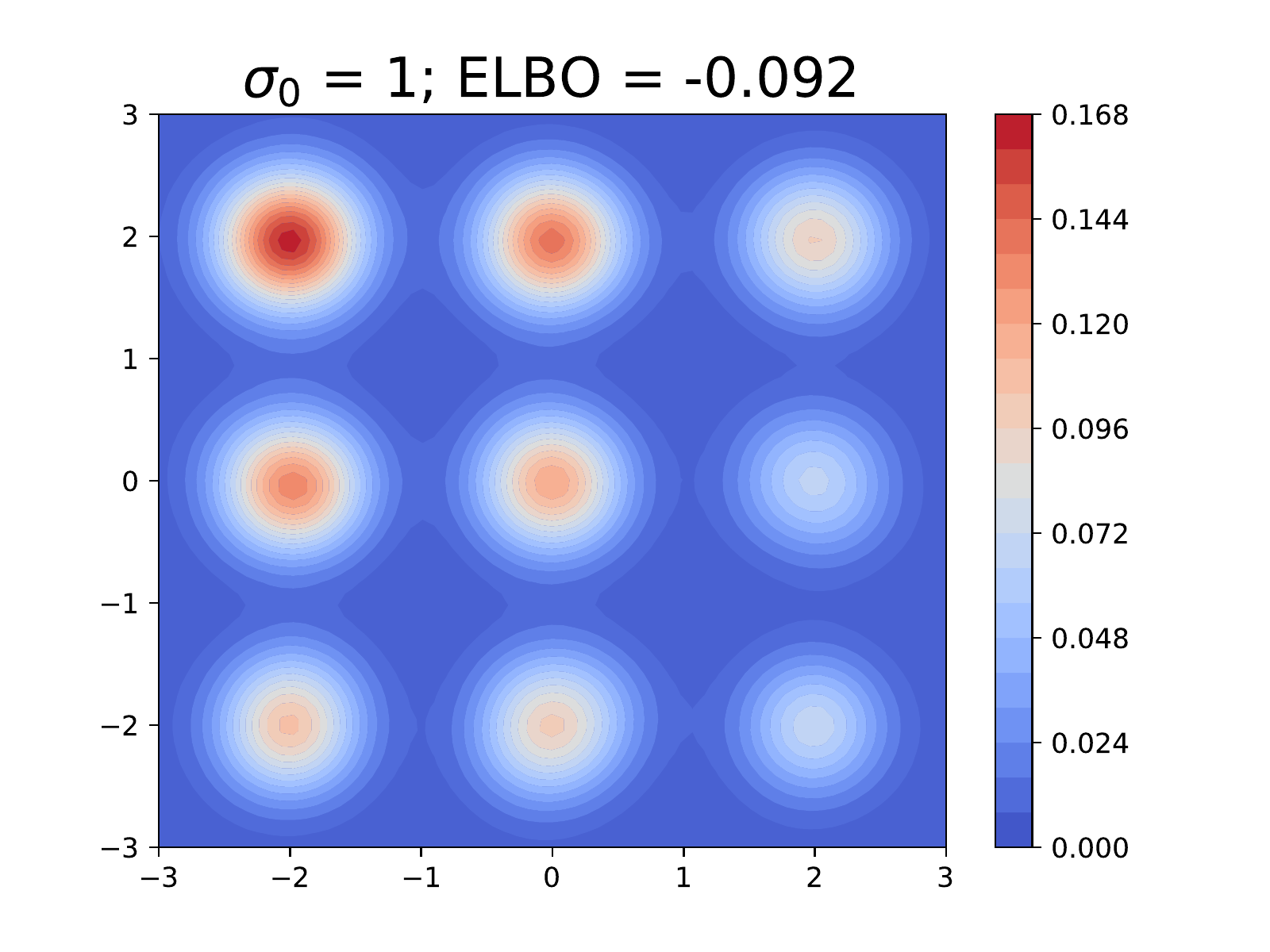}
  \end{subfigure}
\end{minipage}%
\begin{minipage}{.33\textwidth}
  \begin{subfigure}{\linewidth}
    \centering
    \includegraphics[width=0.9\linewidth]{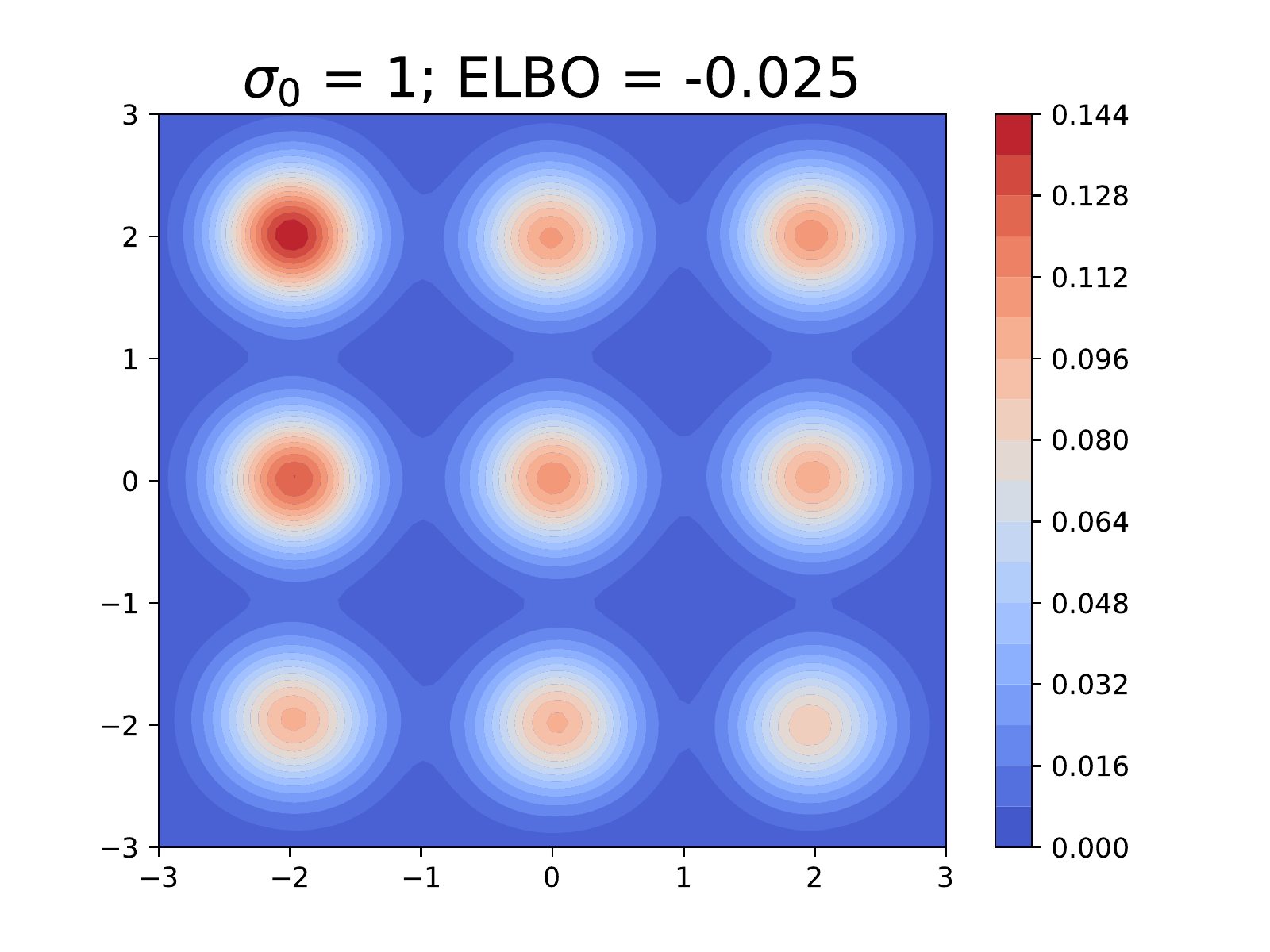}
  \end{subfigure}
  \begin{subfigure}{\linewidth}
    \centering
    \includegraphics[width=0.9\linewidth]{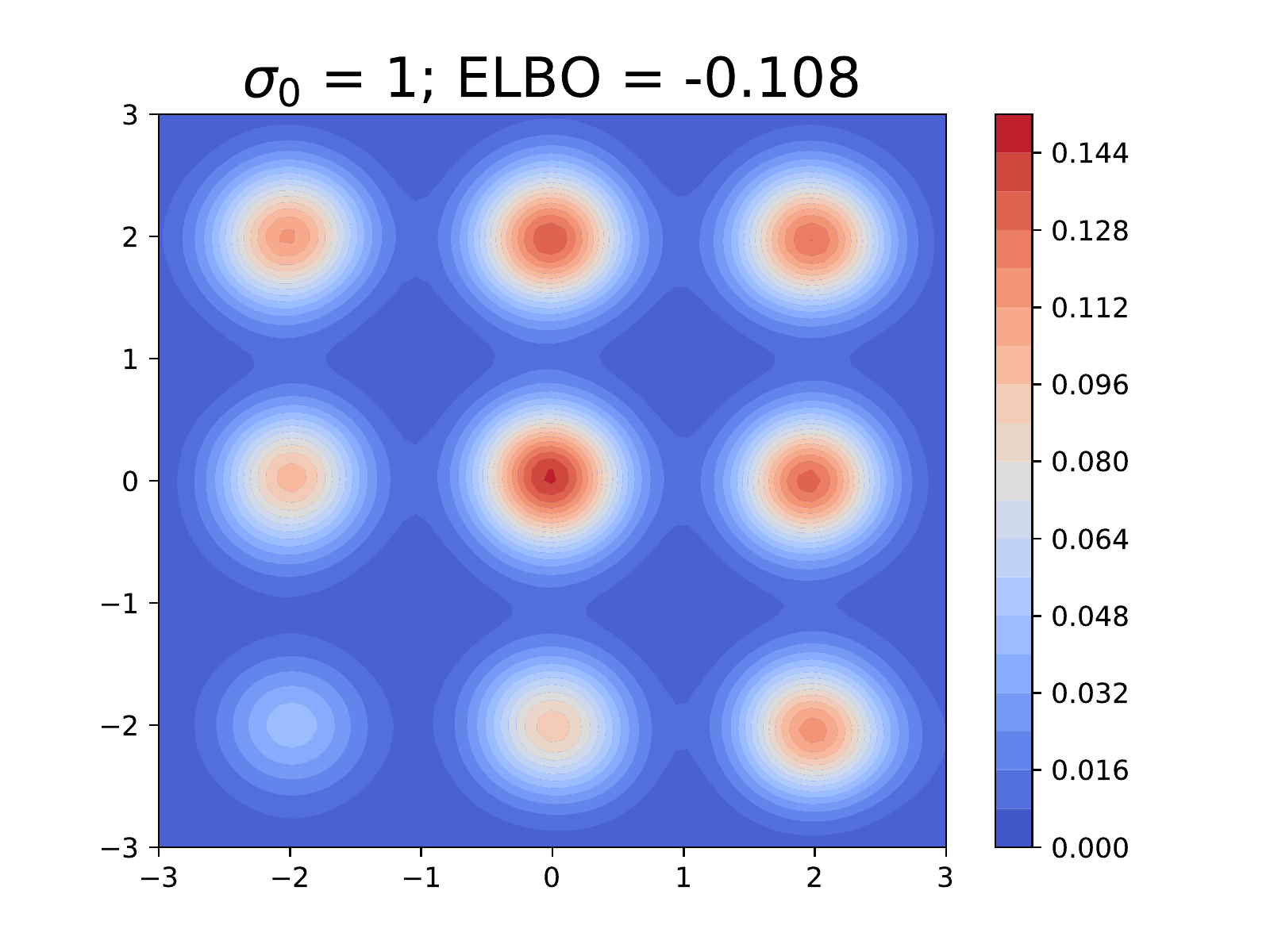}
  \end{subfigure}
\end{minipage}%
\begin{minipage}{.33\textwidth}
  \begin{subfigure}{\linewidth}
    \centering
    \includegraphics[width=0.9\linewidth]{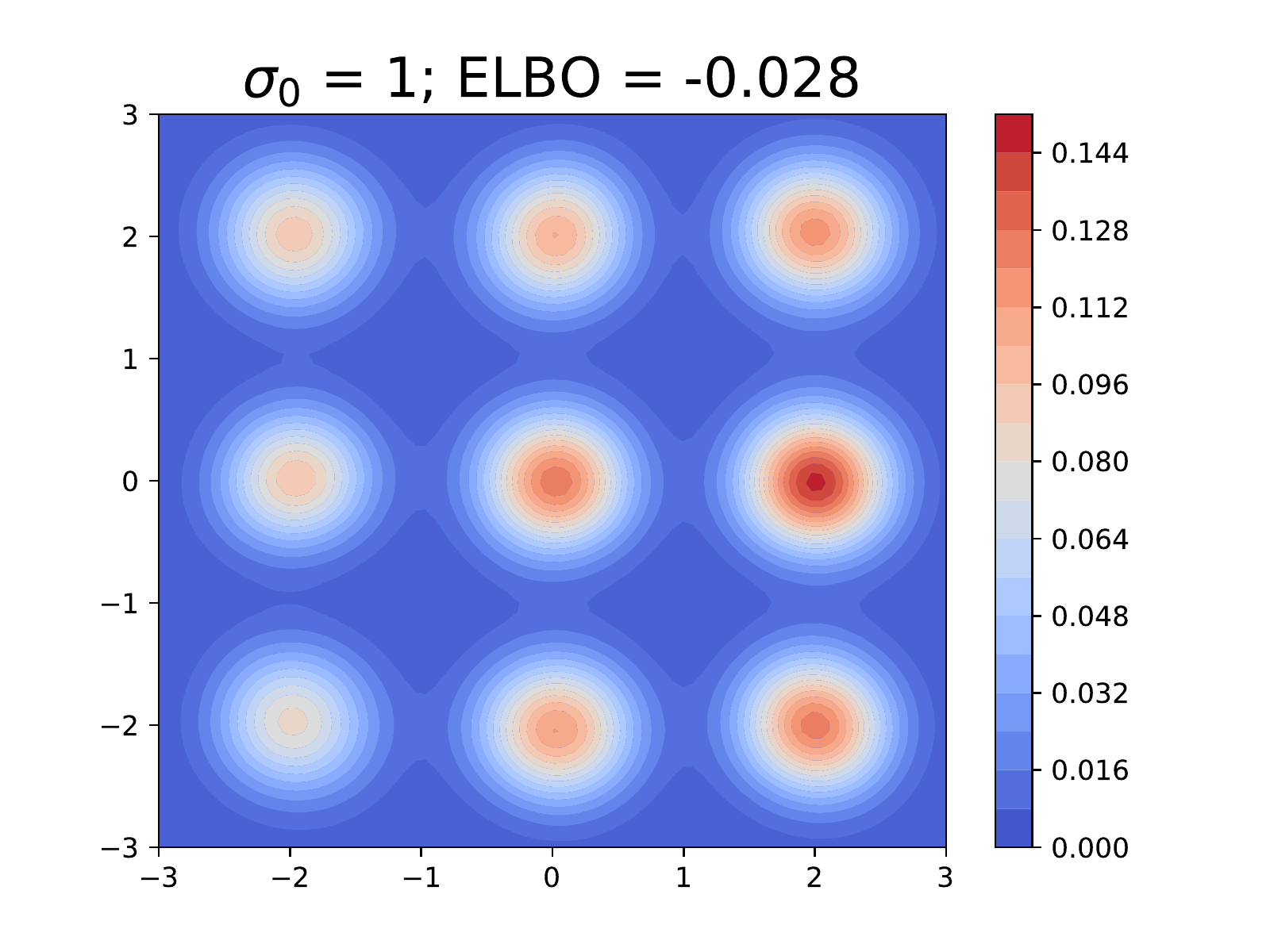}
  \end{subfigure}
  \begin{subfigure}{\linewidth}
    \centering
    \includegraphics[width=0.9\linewidth]{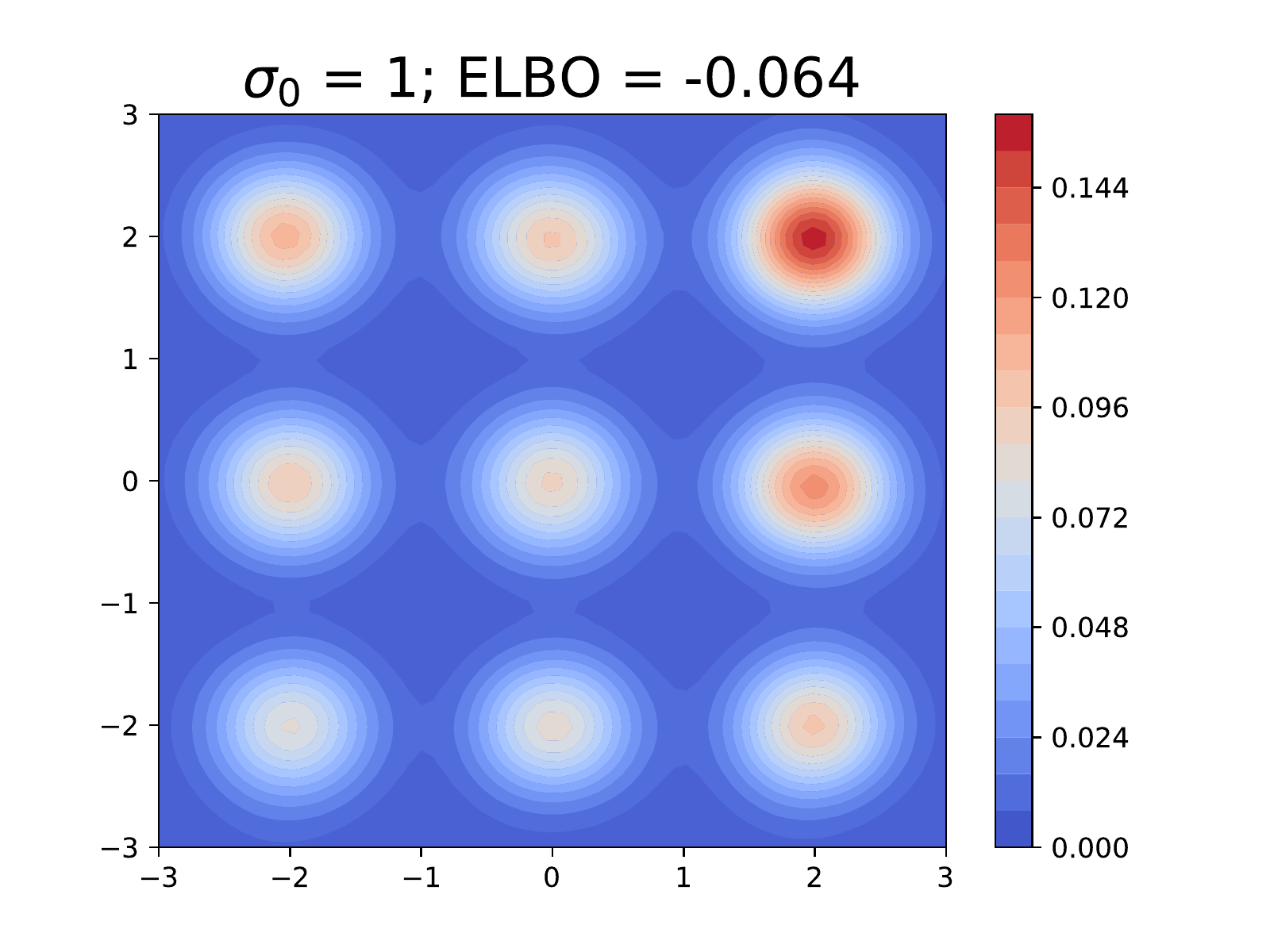}
  \end{subfigure}
\end{minipage}%
\caption{All runs of the mixture of Gaussians experiment for $\sigma_0 = 1$.
CIF models are in the top row, NSF models in the bottom row.}
\label{fig:all-mog-medium-sigma0}
\end{figure*}

\begin{figure*}
\begin{minipage}{.33\textwidth}
  \begin{subfigure}{\linewidth}
    \centering
    \includegraphics[width=0.9\linewidth]{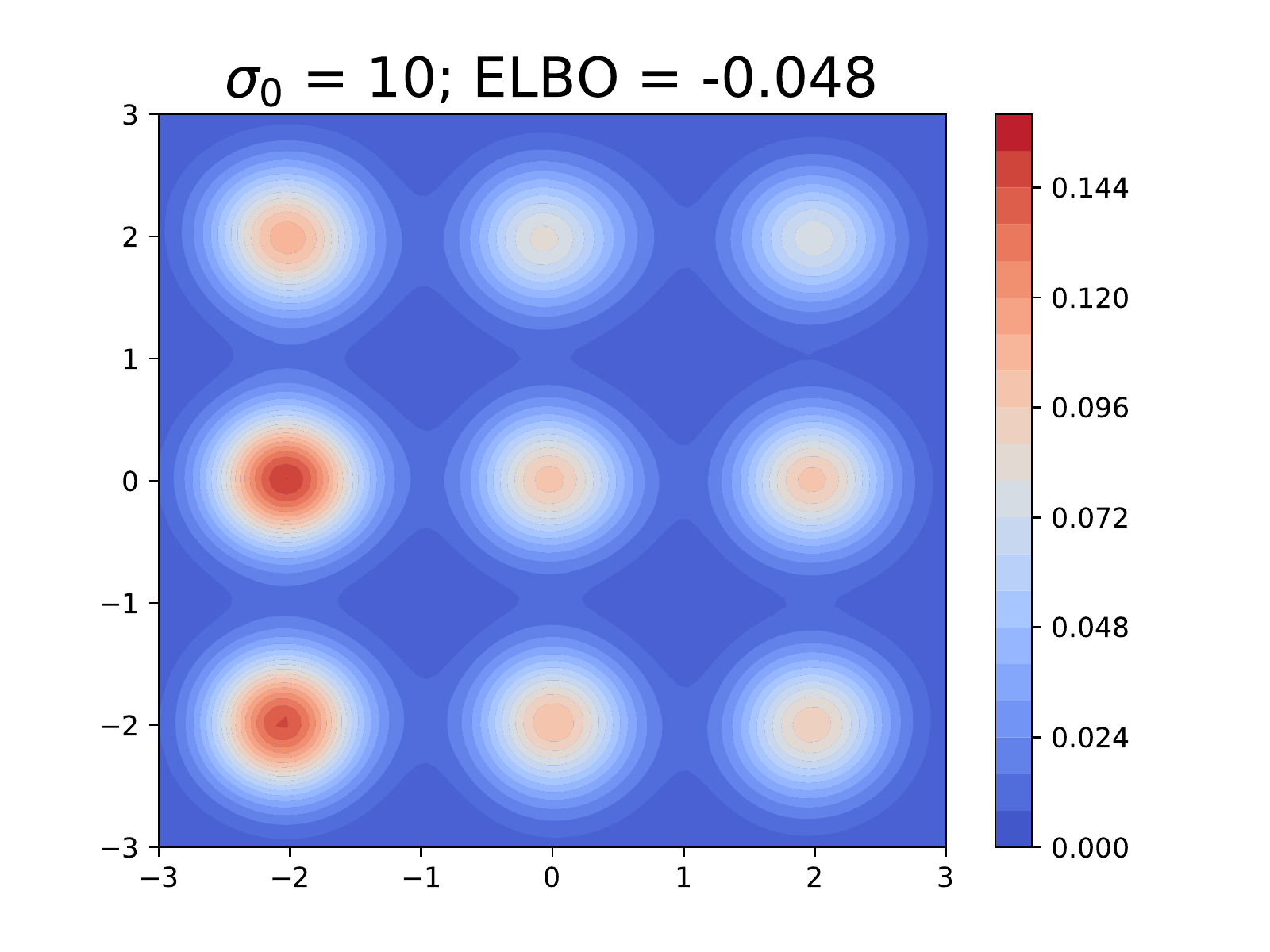}
  \end{subfigure}
  \begin{subfigure}{\linewidth}
    \centering
    \includegraphics[width=0.9\linewidth]{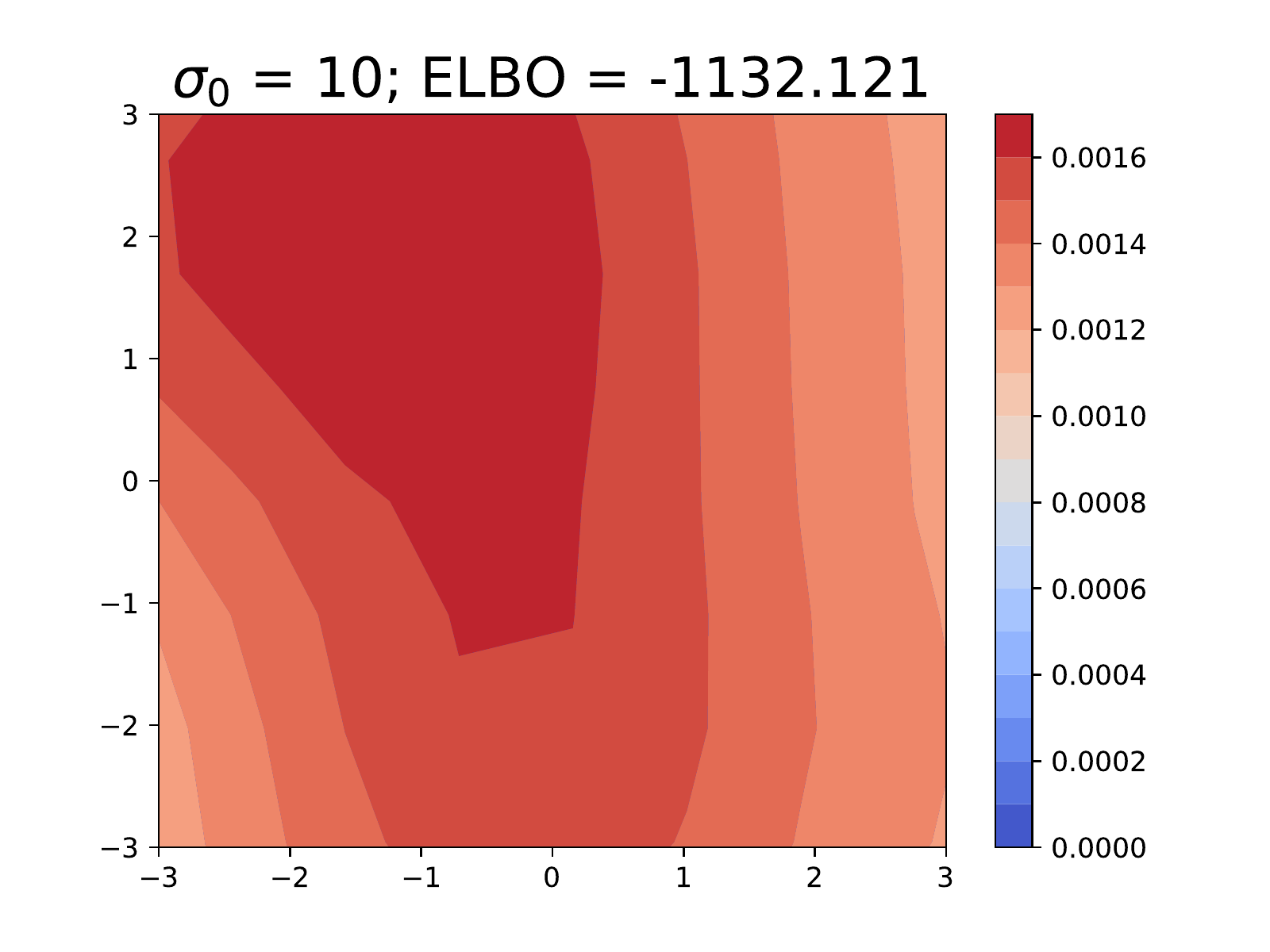}
  \end{subfigure}
\end{minipage}%
\begin{minipage}{.33\textwidth}
  \begin{subfigure}{\linewidth}
    \centering
    \includegraphics[width=0.9\linewidth]{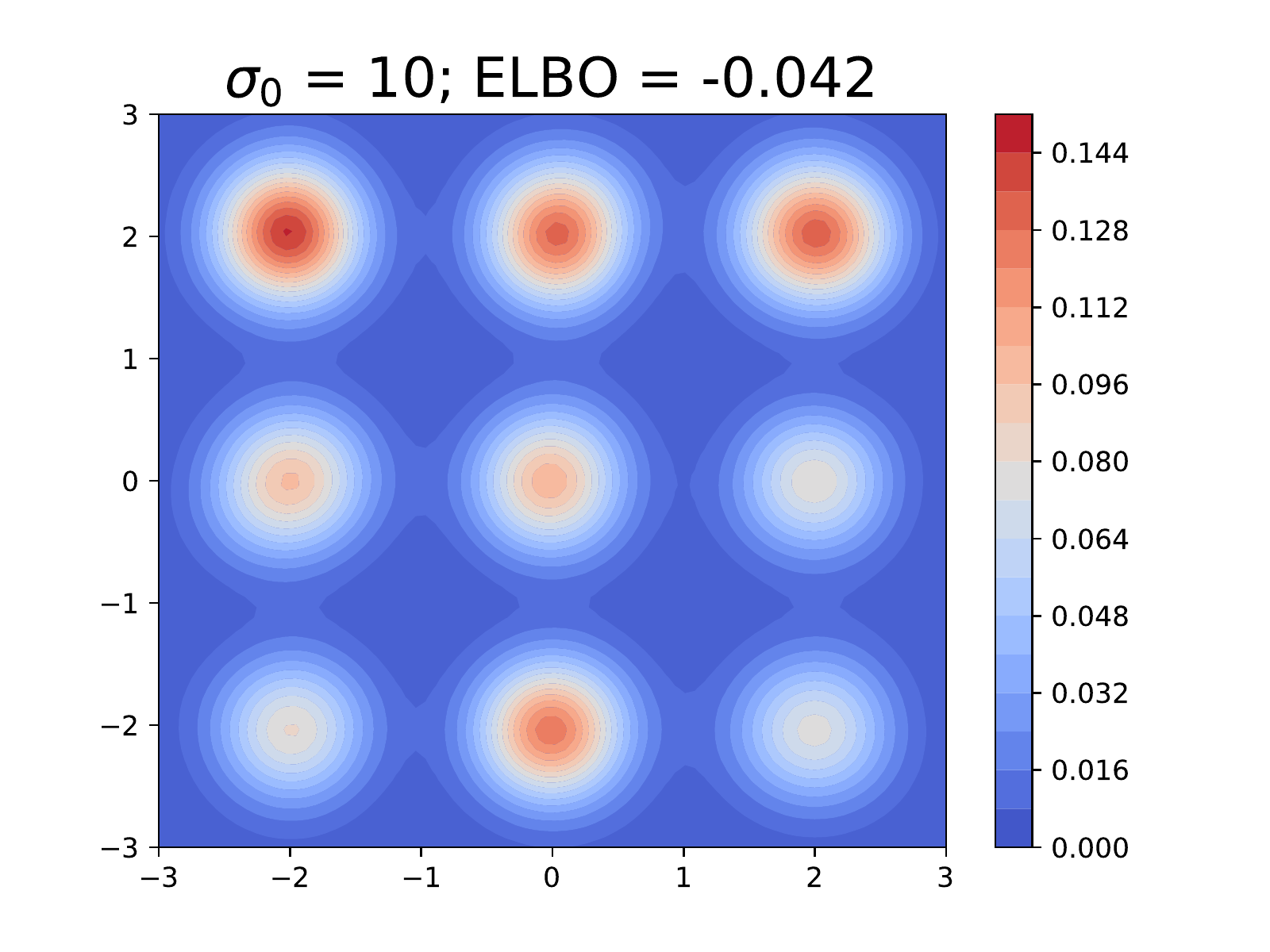}
  \end{subfigure}
  \begin{subfigure}{\linewidth}
    \centering
    \includegraphics[width=0.9\linewidth]{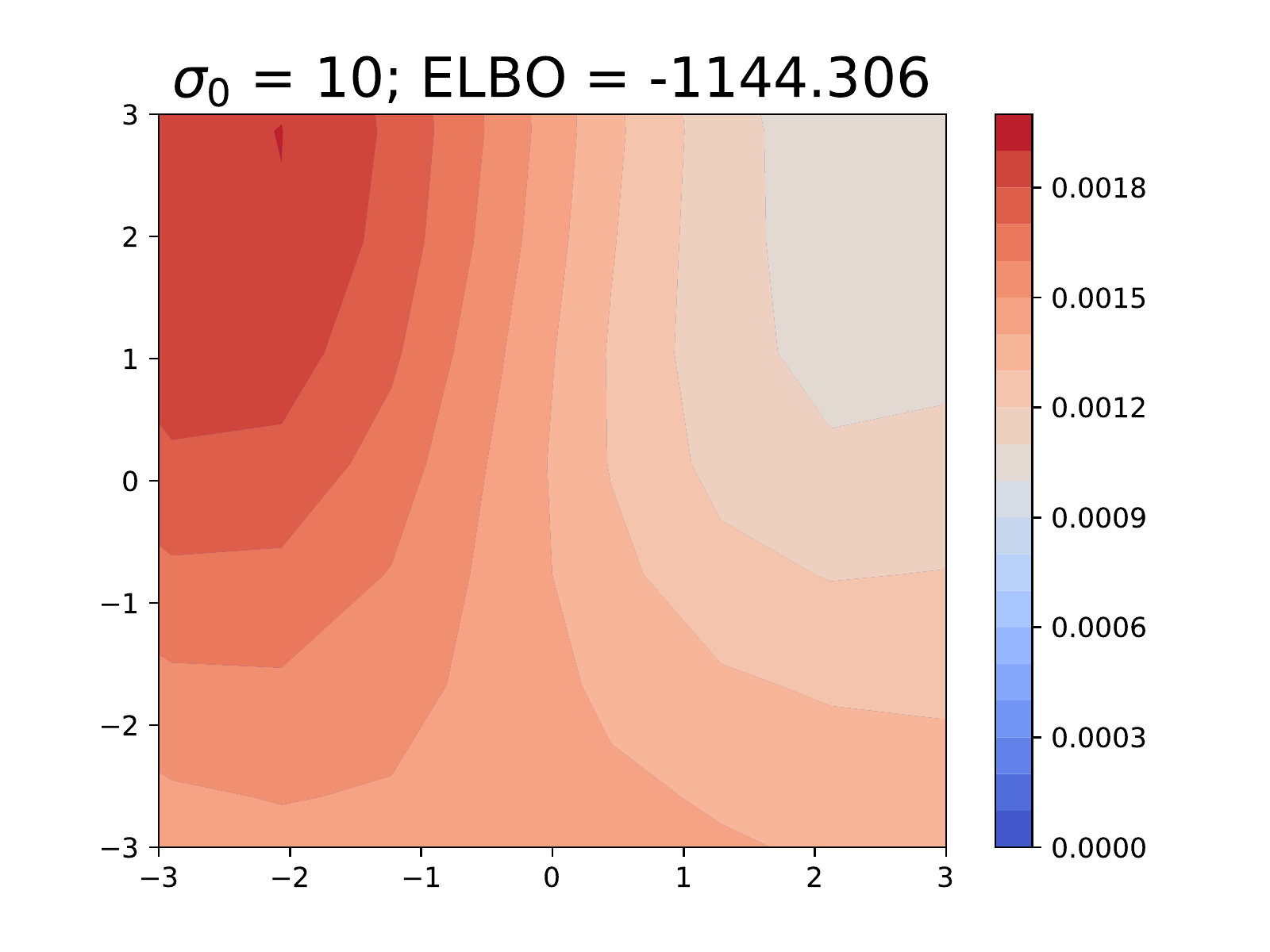}
  \end{subfigure}
\end{minipage}%
\begin{minipage}{.33\textwidth}
  \begin{subfigure}{\linewidth}
    \centering
    \includegraphics[width=0.9\linewidth]{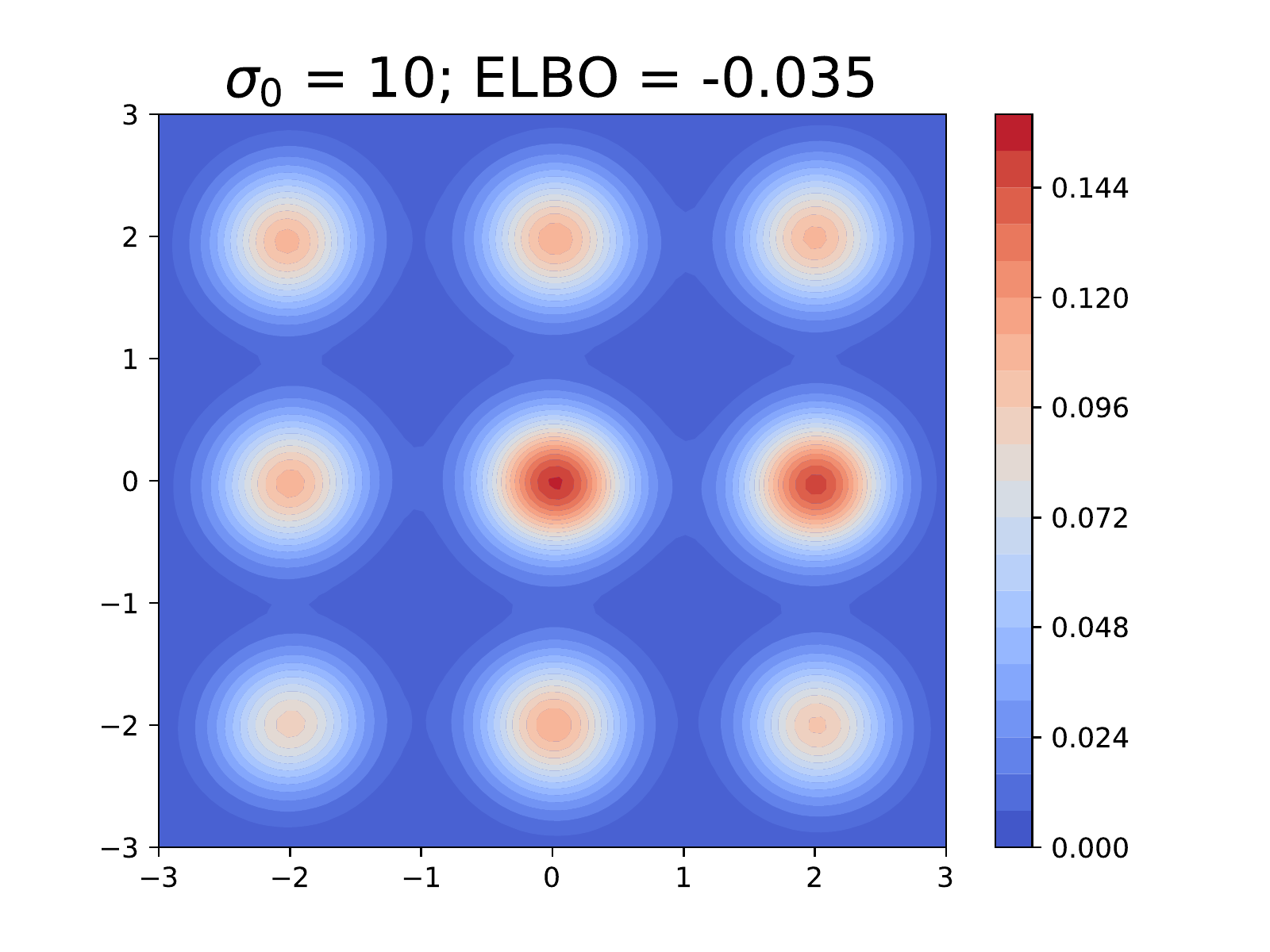}
  \end{subfigure}
  \begin{subfigure}{\linewidth}
    \centering
    \includegraphics[width=0.9\linewidth]{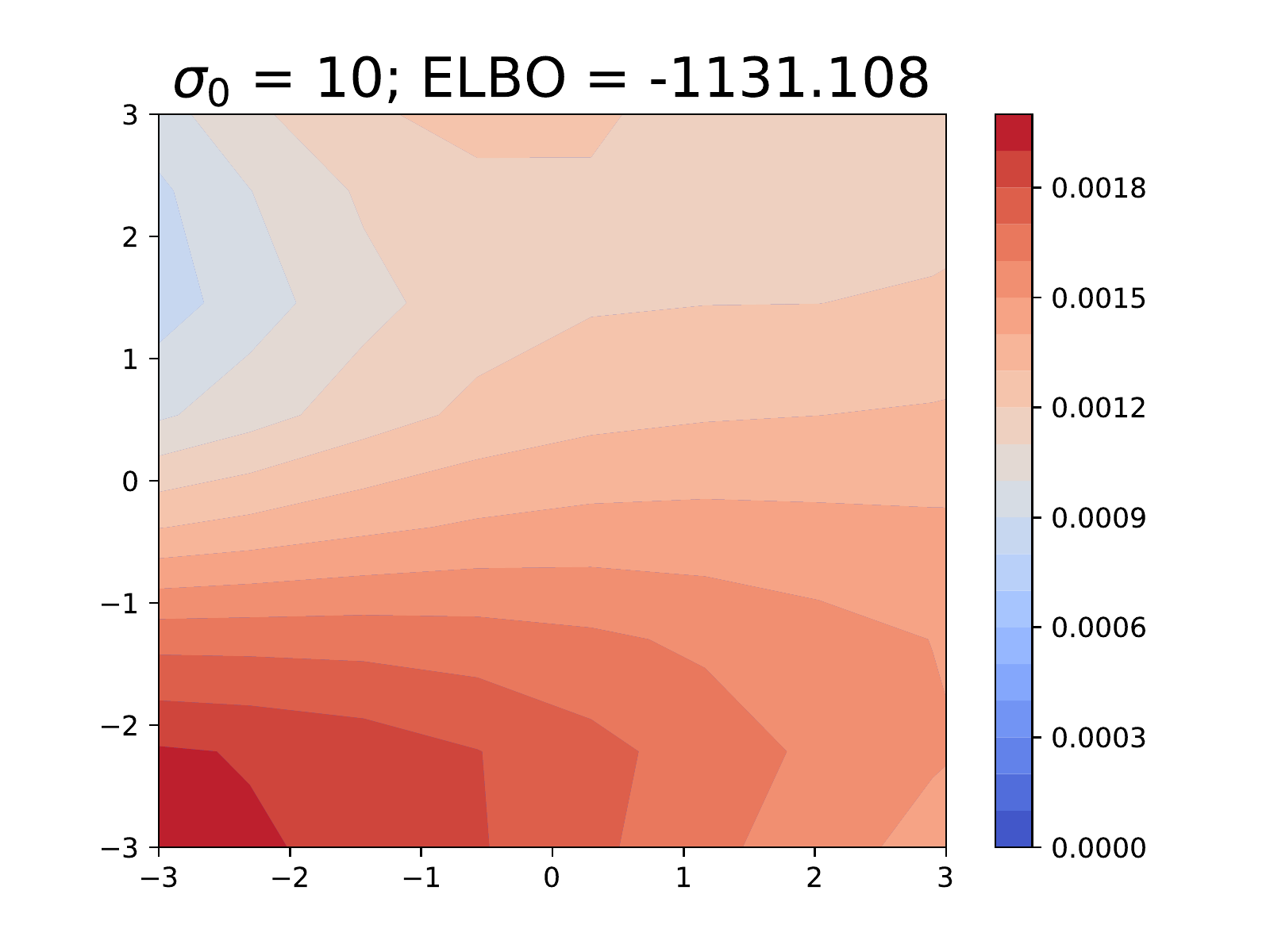}
  \end{subfigure}
\end{minipage}%
\caption{All runs of the mixture of Gaussians experiment for $\sigma_0 = 10$.
CIF models are in the top row, NSF models in the bottom row.}
\label{fig:all-mog-big-sigma0}
\end{figure*}

We have included visualizations of the trained approximate posteriors for all three runs of each setting of $\sigma_0$ in \autoref{fig:all-mog-small-sigma0}, \autoref{fig:all-mog-medium-sigma0}, and \autoref{fig:all-mog-big-sigma0}.
We notice consistently better performance from the trained CIF models, with the CIFs learning to cover the modes in all cases.

\subsection{Setup of Specific Problems}

\subsubsection{Mixture of Gaussians Experiment}

First of all, we note that the Mixture of Gaussians experiment may seem a bit unusual because we directly define the posterior and have no actual ``data'' $x$ in the problem.
However, we can easily imagine a Bayesian generative process which would essentially create such a posterior:
\[
    z \sim \sum_k \alpha_k \cdot \mathcal N(\mu_k, \covmat_k), \qquad x_i \overset{\text{i.i.d.}}{\sim} \mathcal N(z, \covmat).
\]
Then, $p_{Z|X_{1:n}}(\cdot \mid x_{1:n}) = \sum_k \omega_k(x_{1:n}) \cdot \mathcal N(\tilde \mu_k(\bar x), \tilde \covmat_k)$ is another mixture of Gaussians for some weights $\omega_k$ and modified parameters $\tilde \mu_k, \tilde \Sigma_k$.
Instead of defining the model in this way, we just directly specify the posterior as a mixture of Gaussians and perform inference.

For the $K = 9$ experiment, we evenly space the means out in a lattice within the $[-2,2]^2$ square, i.e.\ $\{\mu_k\}_{k=1}^9 \coloneqq \{-2,0,2\} \times \{-2, 0, 2\}$, and we select $\covmat_k \coloneqq \frac{1}{4^2} \mathbf I$ for all $k \in \{1, \ldots, K\}$ so that the components had enough separation.

For the $K = 16$ experiment, we again evenly space the means out in a lattice, but this time within the $[-3, 3]^2$ square, i.e.\ $\{\mu_k\}_{k=1}^{16} \coloneqq \{-3, -1, 1, 3\} \times \{-3, -1, 1, 3\}$, and again set $\covmat_k \coloneqq \frac{1}{4^2} \mathbf I$ for all $k \in \{1, \ldots, K\}$.

\subsubsection{Image Experiment} \label{sec:image_details}

We first re-iterate that we set $n_Z \coloneqq 20$ for the small decoder, and $n_Z \coloneqq 32$ for the large one, so that $\Z \coloneqq \R^{n_Z}$.

The small decoder is a single-hidden-layer transposed convolutional network.
It applies a fully-connected layer with $\tanh$ nonlinearity to transform the $n_Z$-dimensional latent variables into $8$ feature maps of size $14 \times 14$, and then applies a zero-padded transposed convolution with a $4 \times 4$ kernel and stride of $2$ to project into size $1 \times 28 \times 28$ (the same size as the MNIST or Fashion-MNIST data).
We use this output to directly parametrize the logits of a Bernoulli distribution.

The large decoder exactly matches the form from \citet{durkan2019neural}; indeed, we simply used the \texttt{ConvDecoder} directly from their codebase (\url{https://github.com/bayesiains/nsf}).

We use the standard train/test split for both MNIST and Fashion-MNIST, with $60{,}000$ training points and $10{,}000$ test points in each dataset.
Of the $60{,}000$ training points in each, we set aside $10\%$ as validation points for early stopping.

\subsection{Model Details}

Here we discuss the details of the models used.
We note that all of the MAF, NSF, CIF-MAF, and CIF-NSF models use a distribution specified by the VAE encoder (\autoref{sec:vae_encoder}) as the initial distribution $q_{W|X}$ for the image experiments.

\subsubsection{Masked Autoregressive Flow Bijection Settings}

We note the hyperparameter settings that we used for the masked autoregressive flow bijections throughout the paper in \autoref{tab:maf_params}.
We insert batch normalization layers between flow steps in the MAF models as per the recomnmendation of \citet{papamakarios2017masked}, but do not use them in CIFs as the form of $G_\ell$ makes them unnecessary.

\subsubsection{Neural Spline Flow Bijection Settings}

We note the hyperparameter settings that we used for the neural spline flow bijections throughout the paper in \autoref{tab:nsf_params}.
We clip the gradients at a norm of $5$ in all models using NSF bijections as recommended by \citet{durkan2019neural}.

\begin{table}
\caption{Hyperparameters used in the MAF bijections throughout the paper.}
\label{tab:maf_params}
\begin{center}
\begin{tabular}{l|l}
\toprule
\bfseries Hyperparameter & \bfseries Value \\
\midrule
Flow steps          & $5$  \\
Autoregressive networks     & $512 \times 2$ for Large MAF, $420 \times 2$ for all others (including CIF-MAF)  \\
Batch normalization  & \texttt{True} for MAFs, \texttt{False} for CIF-MAFs \\
\bottomrule
\end{tabular}
\end{center}
\end{table}

\begin{table}
\caption{Hyperparameters used in the NSF bijections throughout the paper. Parameters have the same meaning as those from \citet[Table 5]{durkan2019neural}, although we have additionally noted the tail bound used for the splines.}
\label{tab:nsf_params}
\begin{center}
\begin{tabular}{l|l}
\toprule
\bfseries Hyperparameter & \bfseries Value \\
\midrule
Flow steps          & $5$ for Gaussian mixture, $10$ for images  \\
Residual blocks     & $2$  \\
Hidden features     & $44$ for Large NSF, $32$ for all others (including CIF-NSF) \\
Bins                & $8$ \\
Dropout             & $0.0$ \\
Tail bound ($B$)    & $3$ \\
\bottomrule
\end{tabular}
\end{center}
\end{table}

\subsubsection{Continuously-Indexed Flow Settings} \label{sec:cif-settings}

In this section, we describe the network configurations and hyperparameter settings that we use for the CIF extensions to the NSF bijections.
Beyond what is required for the baseline flow, a multi-layer CIF additionally requires definitions of $q_{U_\ell | W_{\ell-1}}$, $r_{U_\ell | W_\ell}$ ($r_{U_\ell | W_\ell, X}$ when amortized), and $s_\ell, t_\ell$ (from \eqref{eq:cif_G}) for $\ell \in \{1, \ldots, L\}$, which we describe below.

For all experiments, we define the densities $q_{U_\ell | W_{\ell-1}}(\cdot \mid w) \coloneqq \mathcal N\left(\mu_\ell^u(w), \text{diag }(\sigma_\ell^u(w)^2)\right)$ for all $\ell \in \{1, \ldots, L\}$ and $w \in \Z$, where $\mu_\ell^u(w)$ and $\sigma_\ell^u(w)$ are outputs of the same neural network: a 2-hidden-layer MLP with $10$ hidden units in each layer.
Similarly, $s_\ell$ and $t_\ell$ are two outputs of a 2-hidden-layer MLP with $10$ hidden units in each layer.

The auxiliary inference model for the Gaussian mixture experiment is essentially the same as $q$ above: $r_{U_\ell|W_\ell}(\cdot \mid w) \coloneqq \mathcal N\left(\mu_\ell^r(w), \text{diag }(\sigma_\ell^r(w)^2)\right)$ for all $\ell \in \{1, \ldots, L\}$ and $w \in \Z$, where $\mu_\ell^r(w)$ and $\sigma_\ell^r(w)$ are outputs of a 2-hidden-layer MLP with $10$ hidden units in each layer.

For the image experiment, the auxiliary inference model is now amortized, with $r_{U_\ell|W_\ell, X}(\cdot \mid w, x) \coloneqq \mathcal N\left(\mu_\ell^r(w, x), \text{diag }(\sigma_\ell^r(w, x)^2)\right)$ for all $\ell \in \{1, \ldots, L\}, w \in \Z$, and $x \in \X$, where $\mu_\ell^r(w, x)$ and $\sigma_\ell^r(w, x)$ are again two outputs of the same neural network.
However, this network has a more complicated structure as it is taking in both vector-valued and image-valued inputs; we describe the steps of the network in the list below:
\begin{enumerate}
    \item Use a linear layer to project $w$ into a shape amenable to upsampling into an image channel (here we selected $1 \times 7 \times 7$ as this shape).
    \item Bilinearly upsample by a factor of $4$ to size $1 \times 28 \times 28$ and append as an additional channel to the input $x$ to get a new input $\tilde x \in \R^{2 \times 28 \times 28}$.
    \item Feed $\tilde x$ into a network of the same form as the VAE encoder in \autoref{sec:vae_encoder}.
\end{enumerate}
The encoder will output the parameters of the normal distribution as required.
We note that the linear layer step could likely be made more parameter-efficient (e.g.\ map to $1 \times 4 \times 4$ and upsample by a factor of $7$), and there are likely other ways to combine vector-valued $w$ and image-valued $x$ more sensibly.
Nevertheless, the design choices made here performed well in practice.

We also need to specify the $u$ dimension for a CIF: we add $u \in \R$ at each layer for the Gaussian mixture example, and $u \in \R^2$ for the image datasets.
This provides a total $u$ dimension of $5$ for the Gaussian mixture example, $10$ for the CIF-MAF, and $20$ for the CIF-NSF.

\subsubsection{VAE Encoder Settings} \label{sec:vae_encoder}

The structure of the encoder used in the VAE model essentially mirrors the structure of the small decoder network from \autoref{sec:image_details}.
In particular, given a $1 \times 28 \times 28$ image, a zero-padded convolution is performed using a $4 \times 4$ filter and stride length $2$ with the $\tanh$ nonlinearity applied afterwards, outputting $8$ feature maps each of size $14 \times 14$.
Then, a fully-connected linear layer is applied to map the feature maps to an output which is two times the size of the latent dimension, giving us the mean and ($\log$) standard deviation of the approximate posterior.

\subsection{Optimization Hyperparameters}

\autoref{tab:opt_params} notes the parameters used for optimizing the models across experiments.
There are a few things to note:
\begin{enumerate}
    \item An ``epoch'' for the mixture of Gaussians example is simple a single stochastic optimization step for a specified number of samples from the approximate posterior since there is no ``data'' in this example.
    \item None of the image experiments actually reached the maximum number of epochs.
    \item The hyperparameter choices below were essentially default choices.
\end{enumerate}

\begin{table}
\caption{Optimization hyperparameters used for each experiment.
Note that an ``epoch'' for the mixture of Gaussians example is simply a single optimization step for a specified number of samples, as there is no ``data''.}
\label{tab:opt_params}
\begin{center}
\begin{tabular}{l|l|l}
\toprule
\bfseries Hyperparameter & \bfseries Mixture of Gaussians & \bfseries Images \\
\midrule
Learning rate               & $10^{-3}$     & $10^{-3}$ \\
Weight decay                & $0$           & $0$       \\
Training batch size         & N/A           & $100$     \\
$q$ samples per step        & $1{,}000$     & $1$       \\
Early stopping              & No            & Yes       \\
Early stopping epochs       & N/A           & $50$      \\
Maximum epochs              & $20{,}000$    & $1{,}000$ \\
\bottomrule
\end{tabular}
\end{center}
\end{table}

\subsection{Estimation of Marginal Log-Likelihood} \label{sec:marg_ll_est}
To generate the log-likelihood outputs in \autoref{tab:images}, we use an importance-sampling-based estimate as in e.g.\ \citet[Appendix~E]{rezende2014stochastic} for each run, and then average the results of this estimator across three runs.
Specifically, given the test dataset $\mathcal D_\text{test} = \{x_i\}_{i=1}^{N_{\text{test}}}$ and a number of samples $S$, the average log-likelihood for a single run is given by
\begin{equation} \label{eq:explicit-log-likelihood}
    \frac{1}{N_\text{test}} \sum_{i=1}^{N_\text{test}} \log \left(  \frac{1}{S} \sum_{s=1}^S\frac{p_{X, Z}(x_i, z_i^{(s)})}{q_{Z|X}(z_i^{(s)} \mid x)} \right),  \qquad \text{where } z_i^{(s)} \sim q_{Z|X}(\cdot \mid x_i),
\end{equation}
for explicit models $q_{Z|X}$ (e.g.\ VAEs and normalizing flows), and
\begin{equation} \label{eq:implicit-log-likelihood}
    \frac{1}{N_\text{test}} \sum_{i=1}^{N_\text{test}} \log \left(  \frac{1}{S} \sum_{s=1}^S\frac{p_{X, Z}(x_i, z_i^{(s)}) \cdot r_{U|Z, X}(u_i^{(s)} \mid z_i^{(s)}, x_i)}{q_{Z, U|X}(z_i^{(s)}, u_i^{(s)} \mid x)} \right),  \qquad \text{where } z_i^{(s)}, u_i^{(s)} \sim q_{Z, U|X}(\cdot, \cdot \mid x_i),
\end{equation}
for implicit models $q_{Z, U | X}$ (e.g.\ CIFs).
We take $S = 1000$ in practice, finding that this provides adequately low-variance estimators as noted in \autoref{tab:estimator_variance}. 

\begin{table}
\caption{Average variance in log-likelihood estimators across models and datasets for the small decoder experiment.
For each run of a particular model on a particular dataset, we calculate the estimator (either \eqref{eq:explicit-log-likelihood} or \eqref{eq:implicit-log-likelihood}) $3$ separate times, and calculate the empirical variance across the outputted estimates.
Then we average this variance across the original $3$ runs for each model-dataset combination, arriving at the numbers in the table.
For example, we have $3$ Small VAE models trained with different random seeds on the MNIST dataset.
For each of these models, we first calculate \eqref{eq:explicit-log-likelihood} three separate times obtaining the empirical variance of these estimates, and then we average the empirical variances across the $3$ Small VAE models trained with different random seeds.
}
\label{tab:estimator_variance}
\begin{center}
\begin{tabular}{l|l|l}
\toprule
\bfseries Model & \bfseries MNIST & \bfseries Fashion-MNIST \\
\midrule
Small VAE   & $2.60 \times 10^{-3}$     & $2.51 \times 10^{-3}$ \\
\midrule
Small MAF   & $2.85 \times 10^{-3}$     & $6.77 \times 10^{-3}$ \\
Large MAF   & $2.59 \times 10^{-3}$     & $4.93 \times 10^{-3}$ \\
CIF-MAF     & $2.38 \times 10^{-3}$     & $6.45 \times 10^{-4}$ \\
\midrule
Small NSF   & $2.78 \times 10^{-4}$     & $3.84 \times 10^{-3}$ \\
Large NSF   & $1.84 \times 10^{-3}$     & $2,27 \times 10^{-3}$ \\
CIF-NSF     & $9.78 \times 10^{-4}$     & $4.47 \times 10^{-3}$ \\
\bottomrule
\end{tabular}
\end{center}
\end{table}
    
\section{Further Details on the Marginal ELBO Estimator} \label{sec:marginal_elbo_estimator}

Here is the full version of the (positively) biased, but still consistent, estimator of \eqref{eq:elbo_2} described in \autoref{sec:mog}:
\begin{equation} \label{eq:marg_elbo_est}
    \widehat \ELBO(x) \coloneqq \frac{1}{N} \sum_{i=1}^N \left( \log p_{X,Z}(x, z_i) - \log \left\{ \frac{1}{M} \sum_{j=1}^M \frac{q_{Z,U}(z_i, u_{i,j})}{r_{U|Z}(u_{i,j} \mid z_i)}  \right\} \right),
\end{equation}
where $(z_i, u'_i) \overset{\text{i.i.d.\ }}{\sim} q_{Z,U}$ for $i \in \{1, \ldots, N\}$, and for each $i$, $u_{i,j} \overset{\text{i.i.d.\ }}{\sim} r_{U|Z}(\cdot \mid z_i)$ for $j \in \{1, \ldots, M\}$.
We need to be careful to make $M$ large enough so that our estimates are not too biased, and at the same time make $N$ large enough so that our estimates are not dominated by variance.
We explore the relationship between values of the estimator \eqref{eq:marg_elbo_est} for a single trained model, and the settings of $M$ and $N$ in \autoref{tab:marginal_elbo_estimator}.
For $N = 10{,}000$ and $M = 100$, we have both acceptably low variance and low bias as the estimator appears to not be significantly lowered by increasing $M$.

\begin{table} \label{tab:estimator}
\caption{Average plus/minus standard deviation of $20$ estimates of the \textbf{marginal} ELBO \eqref{eq:marg_elbo_est} from a single trained model of the mixture of Gaussians experiment.
We vary the values of $N$ and select $M$ as either a linear ($M = N$) or square-root ($M = \sqrt{N}$) function of $N$.
We include Monte Carlo estimates of the auxiliary ELBO from the sample of $z_i$ for reference.
}
\begin{center}
\begin{tabular}{l|ll|ll}
\toprule
\multirow{2}{*}{$N$} &
  \multicolumn{2}{c|}{$M = N$} &
  \multicolumn{2}{c}{$M = \sqrt{N}$} \\
& \bfseries Marginal & \bfseries Auxiliary & \bfseries Marginal & \bfseries Auxiliary \\
\midrule
$1$         & $-0.213 \pm 0.665$    & $-0.237 \pm 0.668$    & $-0.302 \pm 0.361$    & $-0.317 \pm 0.363$ \\
$5$         & $-0.165 \pm 0.180$    & $-0.166 \pm 0.183$    & $-0.178 \pm 0.221$    & $-0.186 \pm 0.223$       \\
$10$        & $-0.189 \pm 0.180$    & $-0.190 \pm 0.187$    & $-0.156 \pm 0.136$    & $-0.160 \pm 0.145$   \\
$50$        & $-0.190 \pm 0.078$    & $-0.198 \pm 0.076$    & $-0.162 \pm 0.094$    & $-0.167 \pm 0.096$   \\
$100$       & $-0.177 \pm 0.057$    & $-0.182 \pm 0.059$    & $-0.181 \pm 0.056$    & $-0.185 \pm 0.053$ \\
$500$       & $-0.176 \pm 0.022$    & $-0.180 \pm 0.022$    & $-0.172 \pm 0.018$    & $-0.177 \pm 0.019$ \\
$1{,}000$   & $-0.160 \pm 0.018$    & $-0.165 \pm 0.019$    & $-0.169 \pm 0.020$    & $-0.174 \pm 0.020$ \\
$5{,}000$   & $-0.169 \pm 0.006$    & $-0.174 \pm 0.006$    & $-0.174 \pm 0.006$    & $-0.179 \pm 0.006$ \\
$10{,}000$  & $-0.171 \pm 0.006$    & $-0.175 \pm 0.006$    & $-0.170 \pm 0.005$    & $-0.175 \pm 0.005$ \\
$50{,}000$  & $-0.169 \pm 0.002$    & $-0.174 \pm 0.002$    & $-0.170 \pm 0.002$    & $-0.175 \pm 0.002$ \\
\bottomrule
\end{tabular}
\end{center}
\label{tab:marginal_elbo_estimator}
\end{table}

\end{document}